\documentclass[10pt,journal,compsoc]{IEEEtran}
\ifCLASSOPTIONcompsoc
  \usepackage[nocompress]{cite}
\else
  \usepackage{cite}
\fi
\usepackage[colorinlistoftodos]{todonotes}
\usepackage{amsmath}
\usepackage{amssymb}
\usepackage{hyperref}

\begin{document}
%
\title{Locate, Size and Count: Accurately Resolving People in Dense Crowds via Detection}
%
%
%
%

\author{Deepak Babu Sam*,
        Skand Vishwanath Peri*,
        Mukuntha Narayanan Sundararaman,
        Amogh Kamath,\\
        R. Venkatesh Babu,~\IEEEmembership{Senior~Member,~IEEE}
\thanks{The authors are with the Video Analytics Lab, Department of Computational and Data Sciences, Indian Institute of Science, Bangalore, India.\protect\\
E-mail: deepaksam@iisc.ac.in, pvskand@pm.me, muks14x@gmail.com, amogh30@gmail.com and venky@iisc.ac.in [* denotes equal contribution]\protect\\
Manuscript received June 8, 2019; revised January 29, 2020.
}
}

\IEEEtitleabstractindextext{%
\begin{abstract}

We introduce a detection framework for dense crowd counting and eliminate the 
need for the prevalent density regression paradigm. Typical counting models 
predict crowd density for an image as opposed to detecting every person. 
These regression methods, in general, fail to localize persons accurate 
enough for most applications other than counting. Hence, we adopt an 
architecture that \textit{locate}s every person in the crowd, \textit{size}s 
the spotted heads with bounding box and then \textit{count}s them. Compared 
to normal object or face detectors, there exist certain unique challenges in 
designing such a detection system. Some of them are direct consequences of the 
huge diversity in dense crowds along with the need to predict boxes contiguously. 
We solve these issues and develop our LSC-CNN model, which can reliably 
detect heads of people across sparse to dense crowds. LSC-CNN employs a 
multi-column architecture with top-down feature modulation to better resolve 
persons and produce refined predictions at multiple resolutions. Interestingly, 
the proposed training regime requires only point head annotation, but can 
estimate approximate size information of heads. We show that LSC-CNN not only 
has superior localization than existing density regressors, but outperforms in 
counting as well. The code for our approach is available at 
{\tt\href{https://github.com/val-iisc/lsc-cnn}{https://github.com/val-iisc/lsc-cnn}}.

\end{abstract}

}

\maketitle

\IEEEdisplaynontitleabstractindextext

%
\IEEEpeerreviewmaketitle

\IEEEraisesectionheading{\section{Introduction}
\label{sect:s1}}

\begin{figure*}[!th]
    \begin{center}            
    \includegraphics[width=\linewidth,height=0.19\textheight]{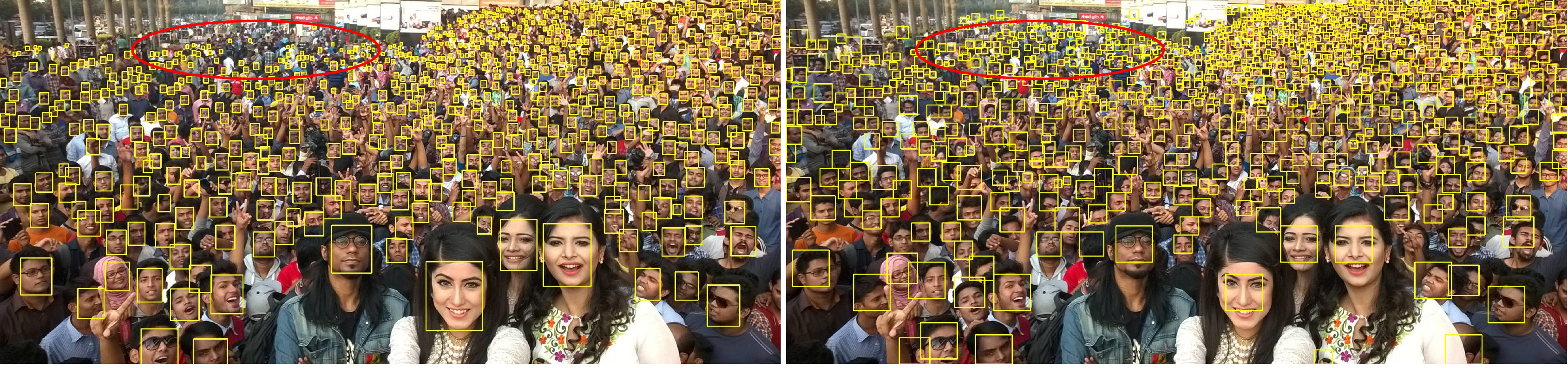}
    \end{center}
    \caption{Face detection vs. Crowd counting. Tiny Face detector \cite{Hu_2017_CVPR}, trained on face dataset \cite{widerface} with box annotations, is able to capture 731 out of the 1151 people in the first image \cite{largestselfi}, losing mainly in highly dense regions. In contrast, despite being trained on  crowd dataset \cite{zhang2016single} having only point head annotations, our LSC-CNN detects 999 persons (second image) consistently across density ranges and provides fairly accurate boxes.}
    \label{fig:f1}
\end{figure*}

\IEEEPARstart{A}{nalyzing} large crowds quickly, is one of the highly sought-after capabilities 
nowadays. Especially in terms of public security and planning, this assumes prime importance. 
But automated reasoning of crowd images or videos is a challenging \emph{Computer Vision} task. 
The difficulty is extreme in dense crowds that the task is typically narrowed down to 
estimating the number of people. Since the count or distribution of people in the scene itself  
can be very valuable information, this field of research has gained traction.

There exists a huge body of works on crowd counting. They range from initial detection 
based methods (\nolinebreak\hspace{1sp}\cite{wu2005detection,viola2005detecting,wang2011automatic,Idrees_tpami}, etc.)  
to later models regressing crowd density 
(\nolinebreak\hspace{1sp}\cite{idrees2013multi,zhang2015cross,zhang2016single,onoro2016towards,sam2017switching,sindagi2017generating,Li_2018_CVPR}, etc.). 
The detection approaches, in general, seem to scale poorly across the entire spectrum of diversity 
evident in typical crowd scenes. Note the crucial difference between the normal face detection 
problem with crowd counting; faces may not be visible for people in all cases (see Figure 
\ref{fig:f1}). In fact, due to extreme pose, scale and view point variations 
, learning a 
consistent feature set to discriminate people seems difficult. Though faces might be largely 
visible in sparse assemblies, people become tiny blobs in highly dense crowds. This makes it 
cumbersome to put bounding boxes in dense crowds, not to mention the sheer number of people, 
in the order of thousands, that need to be annotated per image. Consequently, the problem is 
more conveniently reduced to that of density regression.

In density estimation, a model is trained to map an input image to its crowd density, 
where the spatial values indicate the number of people per unit pixel. To facilitate this, 
the heads of people are annotated, which is much easier than specifying bounding box for 
crowd images \cite{idrees2013multi}. These point annotations are converted to density map by 
convolving with a Gaussian kernel such that simple spatial summation gives out the crowd count. 
Though regression is the dominant paradigm in crowd analysis and delivers excellent count 
estimation, there are some serious limitations. The first being the inability to pinpoint 
persons as these models predict crowd density, which is a regional feature (see the density 
maps in Figure \ref{fig:f2}). Any simple post-processing of density maps to extract positions 
of people, does not seem to scale across the density ranges and results in poor counting performance 
(Section \ref{sect:localization}). Ideally, we expect the model to deliver accurate localization 
on every person in the scene possibly with bounding box. Such a system paves way for downstream 
applications other than predicting just the crowd distribution. With accurate bounding box for 
heads of people in dense crowds, one could do person recognition, tracking etc., which are 
practically more valuable. Hence, we try to go beyond the popular density regression framework and 
create a \emph{dense detection} system for crowd counting.

Basically, our objective is to locate and predict bounding boxes on heads of people, irrespective 
of any kind of variations. Developing such a detection framework is a challenging task and cannot 
be easily achieved with trivial changes to existing detection frameworks 
(\nolinebreak\hspace{1sp}\cite{fasterrcnn,liu2016ssd,Redmon_2017_CVPR,Najibi_2017_ICCV,Hu_2017_CVPR}, etc.). This is because 
of the following reasons: 
\begin{itemize}
\item{\textit{Diversity}:} Any counting model has to handle huge diversity in appearance of 
individual persons and their assemblies. There exist an interplay of multiple variables, 
including but not limited to pose, view-point and illumination variations within the same 
crowd as well as across crowd images.
\item{\textit{Scale}:} The extreme scale and density variations in crowd scenes pose 
unique challenges in formulating a \emph{dense detection} framework. In normal detection scenarios, 
this could be mitigated using a multi-scale architecture, where images are fed to the model at 
different scales and trained. A large face in a sparse crowd is not simply a scaled up version 
of that of persons in dense regions. The pattern of appearance itself is changing across scales 
or density.
\item{\textit{Resolution}:} Usual detection models predict at a down-sampled spatial resolution, 
typically one-sixteenth or one-eighth of the input dimensions. But this approach does not scale across density ranges. 
Especially, highly dense regions require fine grained detection of people, with the possibility 
of hundreds of instances being present in a small region, at a level difficult with the 
conventional frameworks.
\item{\textit{Extreme box sizes}:} Since the densities vary drastically, so should be the 
box sizes. The size of boxes must vary from as small as 1 pixel in highly dense crowds 
to more than 300 pixels in sparser regions, which is several folds beyond the setting under  
which normal detectors operate.
\item{\textit{Data imbalance}:} Another problem due to density variation is the imbalance in box 
sizes for people across dataset. The distribution is so skewed that the majority of samples 
are crowded to certain set of box sizes while only a few are available for the remaining.
\item{\textit{Only point annotation}:} Since only point head annotation is available with crowd 
datasets, bounding boxes are absent for training detectors.
\item{\textit{Local minima}:} Training the model to predict at higher resolutions causes the gradient 
updates to be averaged over a larger spatial area. This, especially with the diverse crowd data, 
increases the chances of optimization being stuck in local minimas, leading to suboptimal 
performance.
\end{itemize}

Hence, we try to tackle these challenges and develop a tailor-made detection framework 
for dense crowd counting. Our objective is to 
\emph{\textbf{L}ocate} every person in the scene, 
\emph{\textbf{S}ize} each detection with bounding box on the head and finally give the crowd 
\emph{\textbf{C}ount}. This LSC-CNN,  
at a functional view, is trained for pixel-wise classification 
task and detects the presence of persons along with the size of the heads. Cross entropy loss 
is used for training instead of the widely employed $l_{2}$ regression loss in density estimation. 
We devise novel solutions to each of the problems listed before, including a method to dynamically 
estimate bounding box sizes from point annotations. 
In summary, this work contributes:
\begin{itemize}
\item \emph{Dense detection} as an alternative to the prevalent density regression paradigm for crowd counting.
\item A novel CNN framework, different from conventional object detectors, 
that provides fine-grained localization of persons at very high resolution.
\item A unique fusion configuration with top-down feature modulation that facilitates 
joint processing of multi-scale information to better resolve people.
\item A practical training regime that only requires point annotations, but 
can estimate boxes for heads.
\item A new winner-take-all based loss formulation for better training at 
higher resolutions.
\item A benchmarked model that delivers impressive performance 
in localization, sizing and counting.
\end{itemize}

\section{Previous Work}

\emph{Person Detection:} The topic of crowd counting broadly might have started 
with the detection of people in crowded scenes. These methods use appearance features 
from still images or motion vectors in videos to detect individual persons 
(\nolinebreak\hspace{1sp}\cite{wu2005detection,viola2005detecting,wang2011automatic}). Idrees et al. 
\cite{Idrees_tpami} leverage local scale prior and global occlusion reasoning to 
detect humans in crowds. With features extracted from a deep network, 
\cite{stewart2015end} run a recurrent framework to sequentially detect and count 
people. In general, the person detection based methods are limited by their 
inability to operate faithfully in highly dense crowds and require bounding box 
annotations for training. Consequently, density regression becomes popular. 

\emph{Density Regression:} Idrees et al. \cite{idrees2013multi} introduce an approach where features 
from head detections, interest points and frequency analysis are used to regress the crowd density. 
A shallow CNN is employed as density regressor in \cite{zhang2015cross}, where the training is done by 
alternatively backpropagating the regression and direct count loss. There are works like 
\cite{wang2015deep}, where the model directly regresses crowd count instead of density map. 
But such methods are shown to perform inferior due to the lack of spatial information in the loss.

\emph{Multiple and Multi-column CNNs:} The next wave of methods focuses on addressing the huge 
diversity of appearance of people in crowd images through multiple networks. Walach et al. 
\cite{walach2016learning} use a cascade of CNN regressors to sequentially correct the errors of 
previous networks. The outputs of multiple networks, each being trained with images of different 
scales, are fused in \cite{onoro2016towards} to generate the final density map. Extending the trend, 
architecture with multiple columns of CNN having different receptive fields starts to emerge. The 
receptive field determines the affinity towards certain density types. For example, the deep network 
in \cite{boominathan2016crowdnet} is supposed to capture sparse crowds, while the shallow network is 
for the blob like people in dense regions. The MCNN \cite{zhang2016single} model leverages three 
networks with filter sizes tuned for different density types. The specialization acquired by individual 
columns in these approaches are improved through a differential training procedure by 
\cite{sam2017switching}. On a similar theme, Sam et al. \cite{Sam_2018_CVPR} create a hierarchical 
tree of expert networks for fine-tuned regression. Going further, the multiple columns are combined 
into a single network, with parallel convolutional blocks of different filters by \cite{Cao_2018_ECCV} and is 
trained along with an additional consistency loss.

\emph{Leveraging context and other information:} Improving density regression by incorporating additional 
information forms another line of thought. Works like (\hspace{1sp}\cite{sindagi2017cnn,sindagi2017generating}) supply 
local or global level auxiliary information through a dedicated classifier. Sam et al. \cite{AAAItdfcnn} show 
a top-down feedback mechanism can effectively provide global context to iteratively improve density prediction 
made by a CNN regressor. A similar incremental density refinement is proposed in \cite{Ranjan_2018_ECCV}.

\emph{Better and easier architectures:} Since density regression suits better for denser crowds, Decide-Net 
architecture from \cite{Liu_2018_CVPR} adaptively leverages predictions from Faster RCNN \cite{fasterrcnn} 
detector in sparse regions to improve performance. Though the predictions seems to be better in sparse crowds, 
the performance on dense datasets is not very evident. Also note that the focus of this work is to aid better 
regression with a detector and is not a person detection model. In fact, Decide-Net requires some bounding box 
annotation for training, which is infeasible for dense crowds. Striving for simpler architectures have always 
been a theme. Li et al. \cite{Li_2018_CVPR} employ a VGG based model with additional dilated convolutional layers 
and obtain better count estimation. Further, a DenseNet \cite{huang2017densely} model is trained in 
\cite{Idrees_2018_ECCV} for density regression at different resolutions with composition loss.

\emph{Other counting works:} An alternative set of works try to incorporate flavours of unsupervised learning and 
mitigate the issue of annotation difficulty. Liu et al. \cite{unlabelled_liu_tpami} use count ranking as a 
self-supervision task on unlabeled data in a multitask framework along with regression supervision on labeled data. 
The Grid Winner-Take-All autoencoder, introduced in \cite{almost_unsup}, trains almost 99\% of the model parameters 
with unlabeled data and the acquired features are shown to be better for density regression. Other counting works 
employ Negative Correlation Learning \cite{Shi_2018_CVPR} and Adversarial training to improve regression 
\cite{Shen_2018_CVPR}. In contrast to all these regression approaches, we move to the paradigm of 
\emph{dense detection}, where the objective is to predict bounding box on heads of people in crowd of any density 
type.

\emph{Object/face Detectors:} Since our model is a detector tailor-made for dense crowds, here we briefly 
compare with other detection works as well. Object detection has seen a shift from early methods relying on 
interest points (like SIFT \cite{Lowe04distinctiveimage}) to CNNs. Early CNN based detectors operate on the 
philosophy of first extracting features from a deep CNN and then have a classifier on the region proposals 
(\hspace{1sp}\cite{girshick2014rich,he2015spatial,girshick2015fast}) or a Region Proposal Network (RPN) \cite{fasterrcnn} 
to jointly predict the presence and boxes for objects. But the current dominant methods 
(\hspace{1sp}\cite{Redmon_2017_CVPR,liu2016ssd}) 
have simpler end-to-end architecture without region proposals. They divide 
the input image in to a grid of cells and boxes are predicted with confidences for each cell. But these works 
generally suit for relatively large objects, with less number instances per image. Hence to capture multiple 
small objects (like faces), many models are proposed. Zhu et al. \cite{zhu2017cms} adapt Faster RCNN 
with multi-scale ROI features to better detect small faces. On similar lines, a pyramid of images 
is leveraged in \cite{Hu_2017_CVPR}, with each scale being separately processed to detect faces of varied sizes. 
The SSH model \cite{Najibi_2017_ICCV} detects faces from multiple scales in a single stage using features from 
different layers. 
More recently, Sindagi et al. \cite{sindagi2019dafe} improves small face detections by enriching 
features with density information.
Our proposed architecture differs from these models in many aspects as described in Section 
\ref{sect:s1}. Though it has some similarity with the SSH model in terms of the single stage architecture, we 
output predictions at resolutions higher than any face detector. This is to handle extremely small heads (of 
few pixels size) occurring very close to each other, a typical characteristic of dense crowds. Moreover, bounding 
box annotation is not available per se from crowd datasets and has to rely on pseudo data. Due to this 
approximated box data, we prefer not to regress or adjust the template box sizes as the normal detectors do, 
instead just classifies every person to one of the predefined boxes. Above all, dense crowd analysis is 
generally considered a harder problem due to the large diversity.

\begin{figure*}[t]
    \begin{center}            
    \includegraphics[width=0.75\linewidth,height=0.21\textheight]{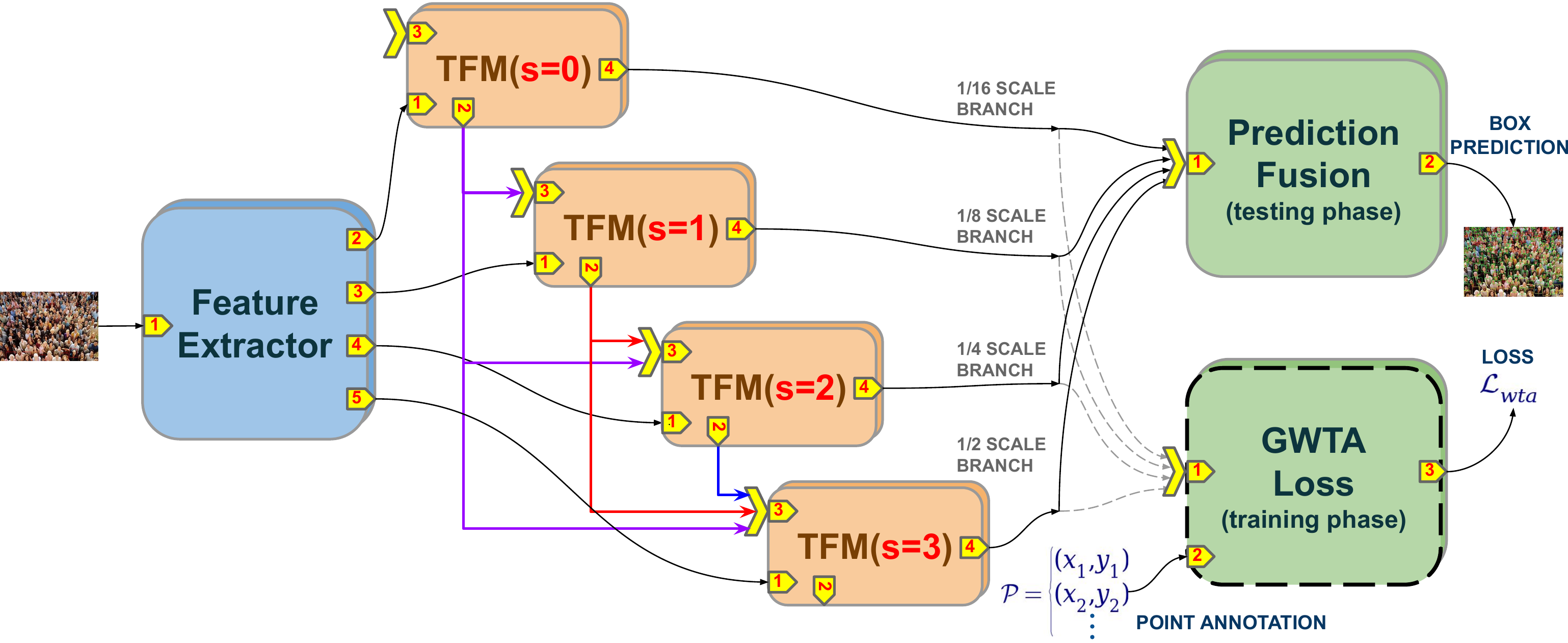}
    \end{center}
    \caption{The architecture of the proposed LSC-CNN is shown. LSC-CNN jointly processes multi-scale information from the feature extractor and provides predictions at multiple resolutions, which are combined to form the final detections. The model is optimized for per-pixel classification of pseudo ground truth boxes generated in the GWTA training phase (indicated with dotted lines).}
    \label{fig:main-arch}
\end{figure*}

\emph{A concurrent work:} We note a recent paper \cite{liu2019point} which proposes a detection framework, 
PSDNN, for crowd counting. But this is a concurrent work which has appeared while this manuscript is under 
preparation. PSDNN uses a Faster RCNN model trained on crowd dataset with pseudo ground truth generated from 
point annotation. A locally constrained regression loss and an iterative ground truth box updation scheme is 
employed to improve performance. Though the idea of generating pseudo ground truth boxes is similar, we do not 
actually create (or store) the annotations, instead a box is chosen from head locations dynamically (Section \ref{sect:size-persons}). We do not regress box location or size as 
normal detectors and avoid any complicated ground truth updation schemes. Also, PSDNN employs Faster RCNN with 
minimal changes, but we use a custom completely end-to-end and single stage architecture tailor-made for the 
nuances of dense crowd detection and outperforms in almost all benchmarks. 

\emph{WTA Architectures}: Since LSC-CNN employs a winner-take-all (WTA) paradigm for training, 
here we briefly compare with similar WTA works. WTA is a biologically inspired widely used case of 
competitive learning in artificial neural networks. In the deep learning scenario, Makhzani et al. 
\cite{makhzani2015winner} propose a WTA regularization for autoencoders, where the basic idea is to 
selectively update only the maximally activated `winner' neurons. This introduces sparsity in weight 
updates and is shown to improve feature learning. The Grid WTA version from \cite{almost_unsup} 
extends the methodology for large scale training and applies WTA sparsity on spatial cells in a 
convolutional output. We follow \cite{almost_unsup} and repurpose the GWTA for supervised training, 
where the objective is to learn better features by restricting gradient updates to the highest loss 
making spatial region (see Section \ref{sect:gwta}).

\section{Our Approach\label{sect:approach}}

As motivated in Section \ref{sect:s1}, we drop the prevalent density regression paradigm 
and develop a \emph{dense detection} model for dense crowd counting. Our model named, 
LSC-CNN, predicts accurately localized boxes on heads of people in crowd images. 
Though it seems like a multi-stage task of first locating and sizing the each 
person, we formulate it as an end-to-end single stage process. Figure \ref{fig:main-arch} 
depicts a high-level view of our architecture. LSC-CNN has three functional parts; the 
first is to extract features at multiple resolution by the \emph{Feature Extractor}. 
These feature maps are fed to a set of \emph{Top-down Feature Modulator} (TFM) networks, 
where information across the scales are fused and box predictions are made. Then 
Non-Maximum Suppression (NMS) selects valid detections from multiple resolutions and 
is combined to generate the final output. For training of the model, the last stage is 
replaced with the \emph{GWTA Loss} module, where the winners-take-all (WTA) loss 
backpropagation and adaptive ground truth box selection are implemented. 
In the following sections, we elaborate on 
each part of LSC-CNN.

\subsection{Locate Heads}

\subsubsection{Feature Extractor}
\label{sect:feat_extr}

\begin{figure}[!b]
    \begin{center}            
    \includegraphics[width=0.7\linewidth,height=0.166\textheight]{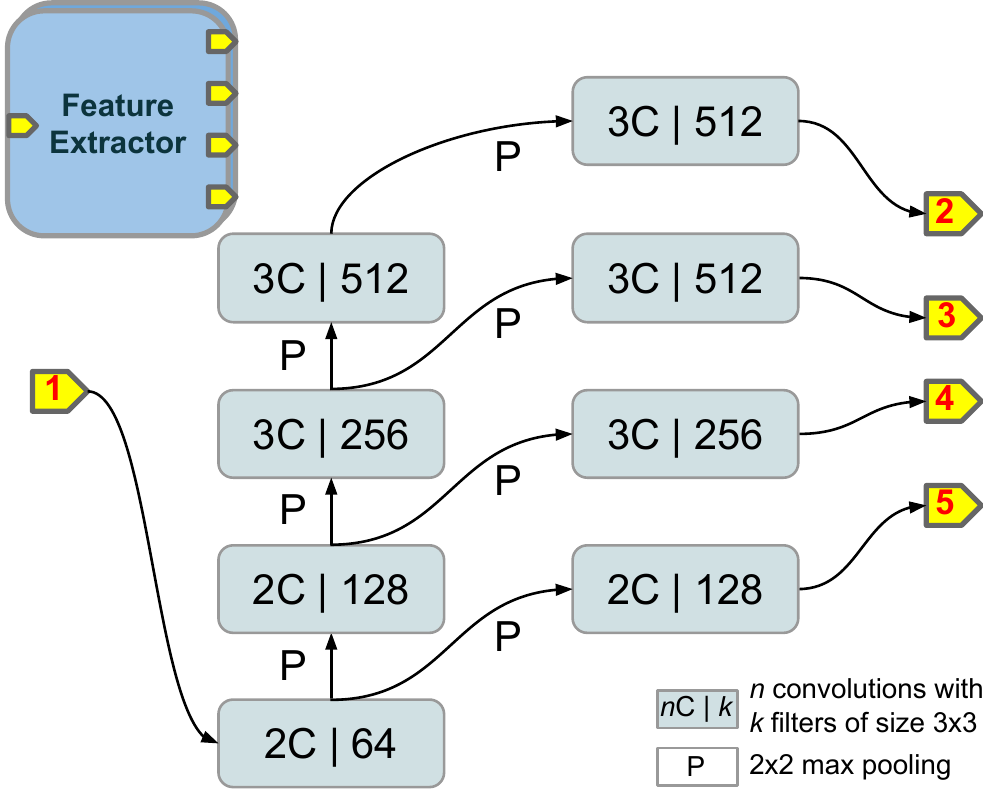}
    \end{center}
    \caption{The exact configuration of \emph{Feature Extractor}, which is a modified version of VGG-16 \cite{simonyan2014very} and outputs feature maps at multiple scales.}
    \label{fig:feature-extractor}
\end{figure}

Almost all existing CNN object detectors operate on a backbone deep feature extractor 
network. The quality of features seems to directly affect the detection performance. 
For crowd counting, VGG-16 \cite{simonyan2014very} based networks are widely used in a 
variety of ways 
(\hspace{1sp}\cite{sam2017switching,Sam_2018_CVPR,sindagi2017generating,sindagi2017cnn}) 
and delivers 
near state-of-the-art performance \cite{Li_2018_CVPR}. In line with the trend, we also 
employ several of VGG-16 convolution layers for better crowd feature extraction. But, as 
shown in Figure \ref{fig:feature-extractor}, some blocks are replicated and manipulated to 
facilitate feature extraction at multiple resolutions. The first five $3\times3$ convolution 
blocks of VGG-16, initialized with ImageNet \cite{ILSVRC15} trained weights, form the backbone 
network. The input to the network is RGB crowd image of fixed size ($224\times224$) with the 
output at each block being downsampled due to max-pooling. At every block, except for the last, 
the network branches with the next block being duplicated (weights are copied at initialization, 
not tied). We tap from these copied blocks to create feature maps at one-half, one-fourth, 
one-eighth and one-sixteenth of the input resolution. This is in slight contrast to typical 
hypercolumn features and helps to specialize each scale branches by sharing low-level features 
without any conflict. The low-level features with half the spatial size that of the input, 
could potentially capture and resolve highly dense crowds. The other lower resolution scale branches 
have progressively higher receptive field and are suitable for relatively less packed ones. In fact, 
people appearing large in very sparse crowds could be faithfully discriminated by the one-sixteenth 
features.

The multi-scale architecture of \emph{feature extractor} is motivated to solve many roadblocks in 
\emph{dense detection}. It could simultaneously address the \emph{diversity}, \emph{scale} and 
\emph{resolution} issues mentioned in Section \ref{sect:s1}. The \emph{diversity} aspect is taken care 
by having multiple scale columns, so that each one can specialize to a different crowd type. Since the 
typical multi-scale input paradigm is replaced with extraction of multi-resolution features, the 
\emph{scale} issue is mitigated to certain extent. Further, the increased resolution for branches 
helps to better resolve people appearing very close to each other.

\subsubsection{Top-down Feature Modulator}
\label{sect:TFM}
One major issue with the multi-scale representations from the feature extractor is that 
the higher resolution feature maps have limited context to discriminate persons. More clearly, 
many patterns in the image formed by leaves of trees, structures of buildings, cluttered 
backgrounds etc. resemble formation of people in highly dense crowds \cite{AAAItdfcnn}. As a 
result, these crowd like patterns could be misclassified as people, especially at the higher 
resolution scales that have low receptive field for making predictions. We cannot avoid these 
low-level representations as it is crucial for resolving people in highly dense crowds. 
The problem mainly arises due to the absence of high-level context information about the crowd 
regions in the scene. Hence, we evaluate global context from scales with higher 
receptive fields and jointly process these top-down features to detect persons.

As shown in Figure \ref{fig:main-arch}, a set of \emph{Top-down Feature Modulator} (TFM) 
modules feed on the output by crowd \emph{feature extractor}. There is one TFM network for 
each scale branch and acts as a person detector at that scale. The TFM also have 
connections from all previous low resolution scale branches. For example, in the case of 
one-fourth branch TFM, it receives connections from one-eighth as well as one-sixteenth 
branches and generates features for one-half scale branch. If there are $s$ feature 
connections from high-level branches, then it uniquely identifies an TFM network as TFM($s$). 
$s$ is also indicative of the scale and takes values from zero to $n_{\mathcal{S}}-1$, where 
$n_{\mathcal{S}}$ is the number of scale branches. For instance, TFM with $s=0$ is for the 
lowest resolution scale (one-sixteenth) and takes no top-down features. Any TFM($s$) with $s>0$ 
receives connections from all TFM($i$) modules where $0\leq i<s$. At a functional view, 
the TFM predicts the presence of a person at every pixel for the given scale branch by 
coalescing all the scale features. This multi-scale feature processing helps to drive global 
context information to all the scale branches and suppress spurious detections, apart from 
aiding scale specialization.

\begin{figure}[t]
    \begin{center}            
    \includegraphics[width=0.7\linewidth, height=0.202\textheight]{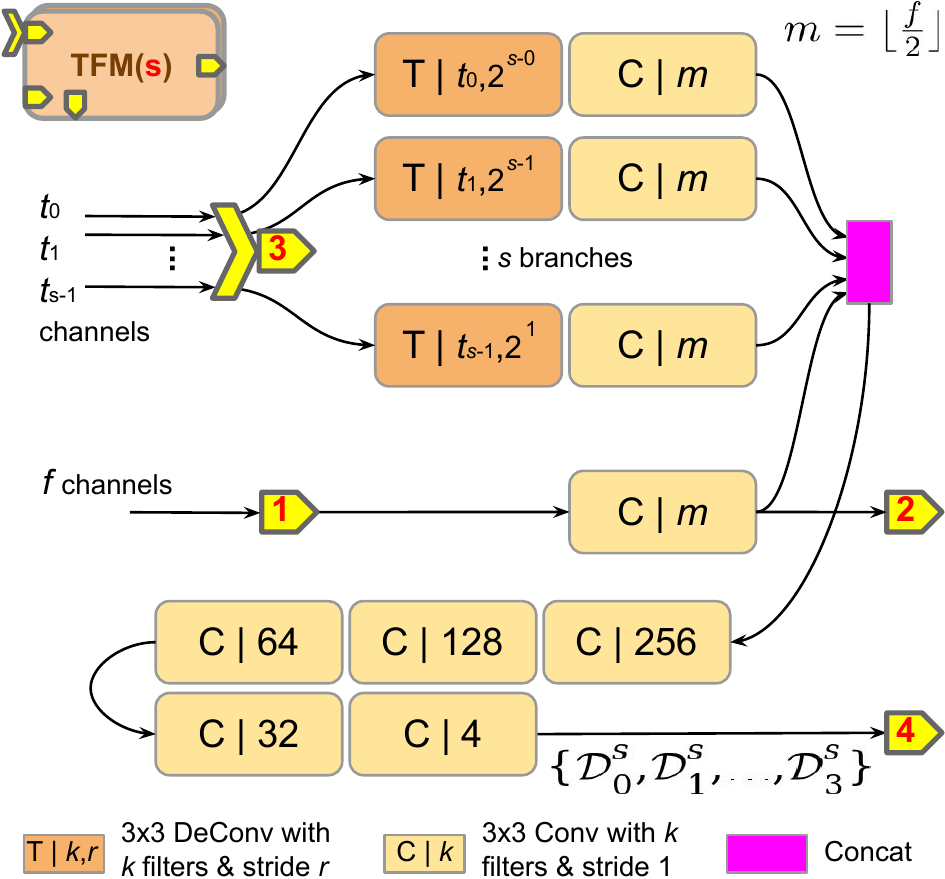}
    \end{center}
    \caption{The implementation of the TFM module is depicted. TFM($s$) processes the features from scale $s$ (terminal $\mathbf{1}$) along with $s$ multi-scale inputs from higher branches (terminal $\mathbf{3}$) to output head detections (terminal $\mathbf{4}$) and the features (terminal $\mathbf{2}$) for the next scale branch.}
    \label{fig:TFM}
\end{figure}

Figure \ref{fig:TFM} illustrates the internal implementation of the TFM module. 
\textbf{Terminal $\mathbf{1}$} of any TFM module takes one of the scale feature 
set from the \emph{feature extractor}, which is then passed through a $3\times3$ 
convolution layer. We set the number of filters for this convolution, $m$, as 
one-half that of the incoming scale branch ($f$ channels from terminal $1$). To be 
specific, $m=\lfloor\frac{f}{2}\rfloor$, where $\lfloor.\rfloor$ denotes floor 
operation. This reduction in feature maps is to accommodate the top-down aggregrated 
multi-scale representations and decrease computational overhead for the final layers. 
Note that the output \textbf{Terminal $\mathbf{2}$} is also drawn from this convolution 
layer and acts as the top-down features for next TFM module. \textbf{Terminal $\mathbf{3}$} 
of TFM($s$) takes $s$ set of these top-down multi-scale feature maps. For the top-down  
processing, each of the $s$ feature inputs is operated by a set of two convolution layers. 
The first layer is a transposed convolution (also know as deconvolution) to upsample top-down 
feature maps to the same size as the scale branch. The upsampling is followed by a 
convolution with $m$ filters. Each processed feature set has the same number of channels 
($m$) as that of the scale input, which forces them to be weighed equally by the subsequent 
layers. All these feature maps are concatenated along the channel dimension and fed to a 
series of $3\times3$ convolutions with progressive reduction in number of filters to give 
the final prediction. These set of layers fuse the crowd features with top-down features 
from other scales to improve discrimination of people. \textbf{Terminal $\mathbf{4}$} 
delivers the output, which basically classifies every pixel into either background or to 
one of the predefined bounding boxes for the detected head. Softmax nonlinearity is 
applied on these output maps to generate per-pixel confidences over the $1+n_{\mathcal{B}}$ 
classes, where $n_{\mathcal{B}}$ is the number of predefined boxes. $n_{\mathcal{B}}$ is a 
hyper-parameter to control the fineness of the sizes and is typically set as 3, making a 
total of $n_{\mathcal{S}}\times n_{\mathcal{B}}=12$ boxes for all the 
branches. The first channel of the prediction for scale $s$, $\mathcal{D}^{s}_{0}$, is for 
background and remaining 
$\{\mathcal{D}^{s}_{1},\mathcal{D}^{s}_{2},\ldots,\mathcal{D}^{s}_{n_{\mathcal{B}}}\}$ 
maps are for the boxes (see Section \ref{sect:box_classif}).

The top-down feature processing architecture helps in fine-grained localization of persons 
spatially as well as in the scale pyramid. 
The \emph{diversity}, \emph{scale} and \emph{resolution} bottlenecks (Section \ref{sect:s1}) are further 
mitigated by the top-down mechanism, which could selectively identify the appropriate scale branch 
for a person 
to resolve it more faithfully. This is further ensured through the training regime we employ (Section 
\ref{sect:gwta}). Scaling across the \emph{extreme box sizes} is also made possible to certain extent 
as each branch could focus on an appropriate subset of the box sizes.

\subsection{Size Heads}
\label{sect:size-persons}

\subsubsection{Box classification}
\label{sect:box_classif}

As described previously, LSC-CNN with the help of TFM modules locates people and 
has to put appropriate bounding boxes on the detected heads. For this sizing, 
we choose a per-pixel classification paradigm. Basically, a set of bounding boxes are 
fixed with predefined sizes and the model simply classifies each head to one of the 
boxes or as background. This is in contrast to the anchor box paradigm typically being 
employed in detectors (\hspace{1sp}\cite{fasterrcnn,Redmon_2017_CVPR}), where the parameters of 
the boxes are regressed. 
Every scale branch ($s$) of the model outputs a set of maps, 
$\{\mathcal{D}^{s}_{b}\}_{b=0}^{n_{\mathcal{B}}}$, indicating the per-pixel confidence for 
the box classes (see Figure \ref{fig:TFM}). Now we require the ground truth sizes of heads to 
train the model, which is not available and not convenient to annotate for typical dense 
crowd datasets. Hence, we devise a method to approximate the sizes of the heads.

For ground truth generation, we rely on the point annotations available with crowd datasets. These 
point annotations specify the locations of heads of people. The location is approximately at 
the center of head, but can vary significantly for sparse crowds (where the point could be any where 
on the large face or head). Apart from locating every person in the crowd, the point annotations 
also give some scale information. For instance, Zhang et al. \cite{zhang2016single} 
use the mean of $k$ nearest neighbours of any head annotation to estimate the adaptive Gaussian 
kernel for creating the ground truth density maps. Similarly, the distance between two adjacent 
persons could indicate the bounding box size for the heads, under the assumption of a smoothly 
varying crowd density. Note that we consider only square boxes. In short, the size for any head 
can simply be taken as the distance to the nearest neighbour. While this approach makes sense in 
medium to dense crowds, it might result in incorrect box sizes for people in sparse crowds, where 
the nearest neighbour is typically far. 
Nevertheless, empirically it is found to work fairly well over a wide range of densities. 

\begin{figure}[!t]
    \begin{center}            
    \includegraphics[width=\linewidth,height=0.1\textheight]{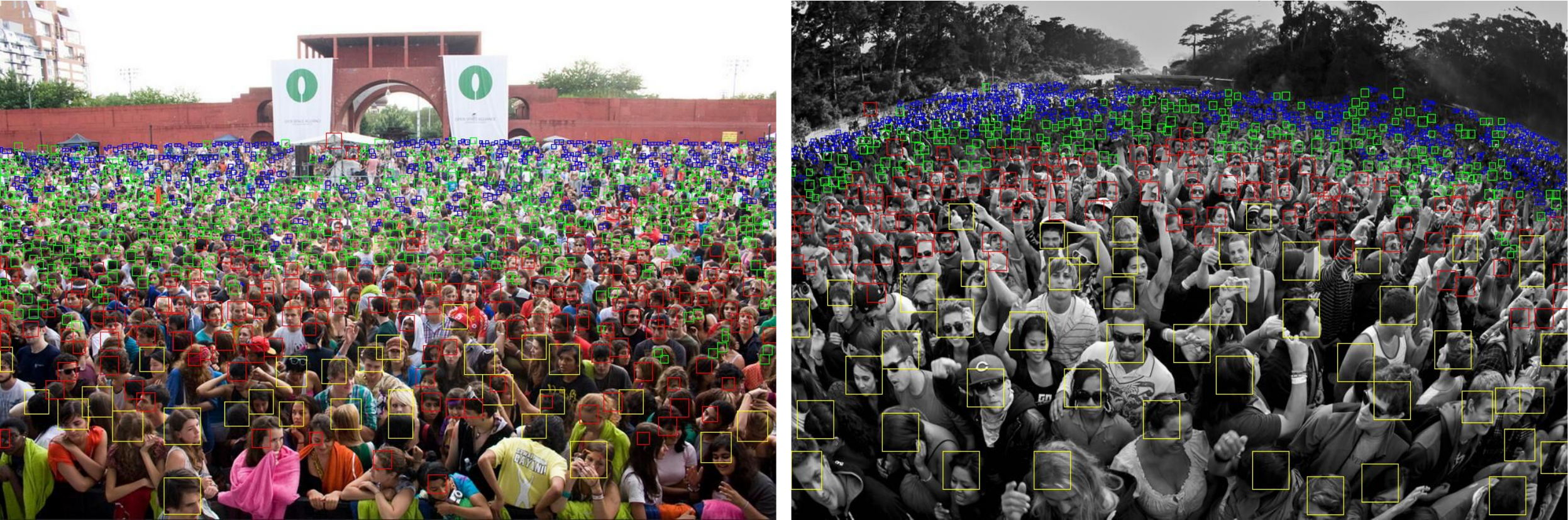}
    \end{center}
    \caption{Samples of generated pseudo box ground truth. Boxes with same color belong to one scale branch.}
    \label{fig:pseudo-gt}
\end{figure}

Here we mathematically explain the pseudo ground truth creation. Let 
$\mathcal{P}$ be the set of all annotated $(x,y)$ locations of people in the given 
image patch. Then for every point $(x,y)$ in $\mathcal{P}$, the box size is defined as, 
\begin{equation}
\mathcal{B}[x,y]=\underset{(x',y')\epsilon P,(x',y')\neq(x,y)}{\textrm{min}}\sqrt{(x-x')^{2}+(y-y')^{2}},
\end{equation}
the distance to the nearest neighbour. If there is only one person in the image patch, the 
box size is taken as $\infty$. Now we discretize the $\mathcal{B}[x,y]$ values to predefined bins, which 
specifies the box sizes. 
Let  $\{\beta^{s}_{1},\beta^{s}_{2},\ldots,\beta^{s}_{n_\mathcal{B}}\}$ be the 
predefined box sizes for scale $s$ and $\overline{\mathcal{B}}_{b}^{s}[x,y]$ denote a boolean 
value indicating whether the location $(x,y)$ belongs to box $b$. Then a person annotation is 
assigned a box $b$ ($\overline{\mathcal{B}}_{b}^{s}[x,y]=1$) if its pseudo size $\mathcal{B}[x,y]$ is between $\beta_{b}^{s}$ and 
$\beta_{b+1}^{s}$. Box sizes less than $\beta^{s}_{1}$ are given to $b=1$ and those greater 
than $\beta^{s}_{n_\mathcal{B}}$ falls to $b=n_\mathcal{B}$. Note that non-people locations 
are assigned $b=0$ background class. 
A general philosophy is followed in choosing the box sizes $\beta^{s}_{b}$s for all the scales. 
The size of the first box ($b=1$) at the highest resolution scale ($s=n_{\mathcal{S}}-1$) is always 
fixed to one, which improves the resolving capacity for highly dense crowds 
(\emph{Resolution} issue in Section \ref{sect:s1}). We choose larger sizes for the remaining boxes 
in the same scale with a constant increment. This increment is fine-grained in higher resolution 
branches, but the coarseness progressively increases for low resolution scales. To be specific, if 
$\gamma^{s}$ represent the size increment for scale $s$, then box sizes are, 
\begin{equation}
\label{equ:pseudo-box-size}
\beta_{b}^{s}=\begin{cases}
\beta_{n_{\mathcal{B}}}^{s+1}+b\gamma^{s} & \textrm{if }s<n_{\mathcal{S}}-1\\
1+(b-1)\gamma^{s} & \textrm{otherwise.}
\end{cases}
\end{equation}
The typical values of the size increment for different scales are $\gamma=\{4,2,1,1\}$. Note that 
the high resolution branches (one-half \& one-fourth) have boxes with finer sizes than the low resolution 
ones (one-sixteenth \& one-eighth), where coarse resolving capacity would suffice (see Figure \ref{fig:pseudo-gt}).

There are many reasons to discretize the head sizes and classify the boxes instead of regressing 
size values. The first is due to the use of pseudo ground truth. Since the size of heads itself is 
approximate, tight estimation of sizes proves to be difficult (see Section \ref{sect:arch-ablations}). 
Similar sized heads in two images could have different ground truths depending on the density. This 
might lead to some inconsistencies in training and could result in suboptimal performance. Moreover, 
the sizes of heads vary extremely across density ranges at a level not expected for normal detectors. 
This requires heavy normalization of value ranges along with complex data balancing schemes. But our 
per-pixel box classification paradigm effectively addresses these \emph{extreme box sizes} and 
\emph{only point annotation} issues (Section \ref{sect:s1}).

\subsubsection{GWTA Training}
\label{sect:gwta}

\textbf{Loss:} We train the LSC-CNN by back-propagating per-pixel cross entropy loss. The loss for a pixel is defined as,
\begin{equation}
l(\{d_{i}\}_{i=0}^{n_{\mathcal{B}}},\{b_{i}\}_{i=0}^{n_{\mathcal{B}}},\{\alpha_{i}\}_{i=0}^{n_{\mathcal{B}}})=-\underset{i=0}{\stackrel{n_{\mathcal{B}}}{\sum}}\alpha_{i}b_{i}\textrm{log}d_{i},
\end{equation}
where $\{d_{i}\}_{i=0}^{n_{\mathcal{B}}}$ is the set of $n_{\mathcal{B}}+1$ probability values (softmax outputs) for the 
predefined box classes and $\{b_{i}\}_{i=0}^{n_{\mathcal{B}}}$ refers to corresponding ground truth labels. All $b_{i}$s 
take zero value except for the correct class. The $\alpha_{i}$s are weights to class balance the training. Now the 
loss for the entire prediction of scale branch $s$ would be,
\[	
\mathcal{L}(\{\mathcal{D}_{b}^{s}\},\{\overline{\mathcal{B}}_{b}^{s}\},\{\alpha_{b}^{s}\})=\underset{x,y}{\overset{w^{s},h^{s}}{\sum}}\frac{l(\{\mathcal{D}_{b}^{s}[x,y]\},\{\overline{\mathcal{B}}_{b}^{s}[x,y]\},\{\alpha_{b}^{s}\})}{w^{s}h^{s}}
\]
where the inputs are the set of predictions $\{\mathcal{D}_{b}^{s}\}_{b=0}^{n_{\mathcal{B}}}$ and pseudo ground truths  
$\{\overline{\mathcal{B}}_{b}^{s}\}_{b=0}^{n_{\mathcal{B}}}$ (the set limits might be dropped for convenience). 
Note that $(w^{s},h^{s})$ are the spatial sizes of these prediction maps and the cross-entropy loss is averaged over it. 
The final loss for LSC-CNN after combining losses from all the branches is,
\begin{equation}
\label{equ:lcomb}
\mathcal{L}_{comb}=\nolinebreak\sum_{s=1}^{n_{\mathcal{S}}}\mathcal{L}(\{\mathcal{D}_{b}^{s}\},\{\overline{\mathcal{B}}_{b}^{s}\},\{\alpha^{s}_{b}\}).
\end{equation}

\textbf{Weighting:} As mentioned in Section \ref{sect:s1}, the \emph{data imbalance} issue is severe in the case of 
crowd datasets. Class wise weighting assumes prime importance for effective backpropagation of $\mathcal{L}_{comb}$ (see Section 
\ref{sect:arch-ablations}). We follow a simple formulation to fix the $\alpha$ values. Once the box sizes are set, 
the number of data points available for each class is computed from the entire train set. Let $c^{s}_{b}$ denote this 
frequency count for the box $b$ in scale $s$. Then for every scale branch, we sum the box counts as 
$c_{sum}^{s}=\sum_{b=1}^{n_{\mathcal{B}}}c^{s}_{b}$ and the scale with minimum number of samples is 
identified. This minimum value $c_{min}=\textrm{min}_{s}c_{sum}^{s}$, is used to balance the training data across 
the branches. Basically, we scale down the weights for the branches with higher counts such that the minimum count branch 
has weight one. Note that training points for all the branches as well as the classes within a branch need to be balanced. 
Usually the data samples would be highly skewed towards the background class ($b=0$) in all the scales. To mitigate this, 
we scale up the weights of all box classes based on its ratio with background frequency of the same branch. Numerically, 
the balancing is done jointly as,
\begin{equation}
\label{equ:weights}
\alpha^{s}_{b}=\frac{c_{min}}{c^{s}_{sum}}\textrm{min}(\frac{c^{s}_{0}}{c^{s}_{b}},10).
\end{equation}
The term $c^{s}_{0}/c^{s}_{b}$ can be large since the frequency of background to box is usually skewed. So we limit the 
value to 10 for better training stability. Further note that for some box size settings, $\alpha^{s}_{b}$ values itself 
could be very skewed, which depends on the distribution of dataset under consideration. Any difference in the values more 
than an order of magnitude is found to be affecting the proper training. Hence, the box size increments ($\gamma$s) are 
chosen not only to roughly cover the density ranges in the dataset, but also such that the $\alpha^{s}_{b}$s are close within 
an order of magnitude.

\textbf{GWTA:} However, even after this balancing, training LSC-CNN by optimizing joint loss $\mathcal{L}_{comb}$ does not 
achieve acceptable performance (see Section \ref{sect:arch-ablations}). This is because the model predicts at a high 
resolution than any typical crowd counting network and the loss is averaged over a relatively larger spatial area. 
The weighing scheme only makes sure that the averaged loss values across branches and classes is in similar range. But the 
scales with larger spatial area could have more instances of one particular class than others. For instance in dense 
regions, the one-half resolution scale ($s=3$) would have more person instances and are typically very diverse. This 
causes the optimization to focus on all instances equally and might lead to a local minima solution. A strategy is 
needed to focus on a small region at a time, update the weights and repeat this for another region.

For solving this \emph{local minima} issue (Section \ref{sect:s1}), we rely on the Grid Winner-Take-All (GWTA) 
approach introduced in \cite{almost_unsup}. Though GWTA is originally used for unsupervised learning, we 
repurpose it to our loss formulation. The basic idea is to divide the prediction map into a grid of cells 
with fixed size and compute the loss only for one cell. Since only a small region is included in the loss, 
this acts as tie breaker and avoids the gradient averaging effect, reducing the chances of the optimization 
reaching a local minima. Now the question is how to select the cells. The `winner' cell is chosen as the one 
which incurs the highest loss. At every iteration of training, we concentrate more on the high loss making 
regions in the image and learn better features. This has slight resemblance to hard mining approaches, where 
difficult instances are sampled more during training. In short, GWTA training selects `hard' regions and try 
to improve the prediction (see Section \ref{sect:arch-ablations} for ablations).

\begin{figure}[t]
    \begin{center}            
    \includegraphics[width=0.89\linewidth, height=0.169\textheight]{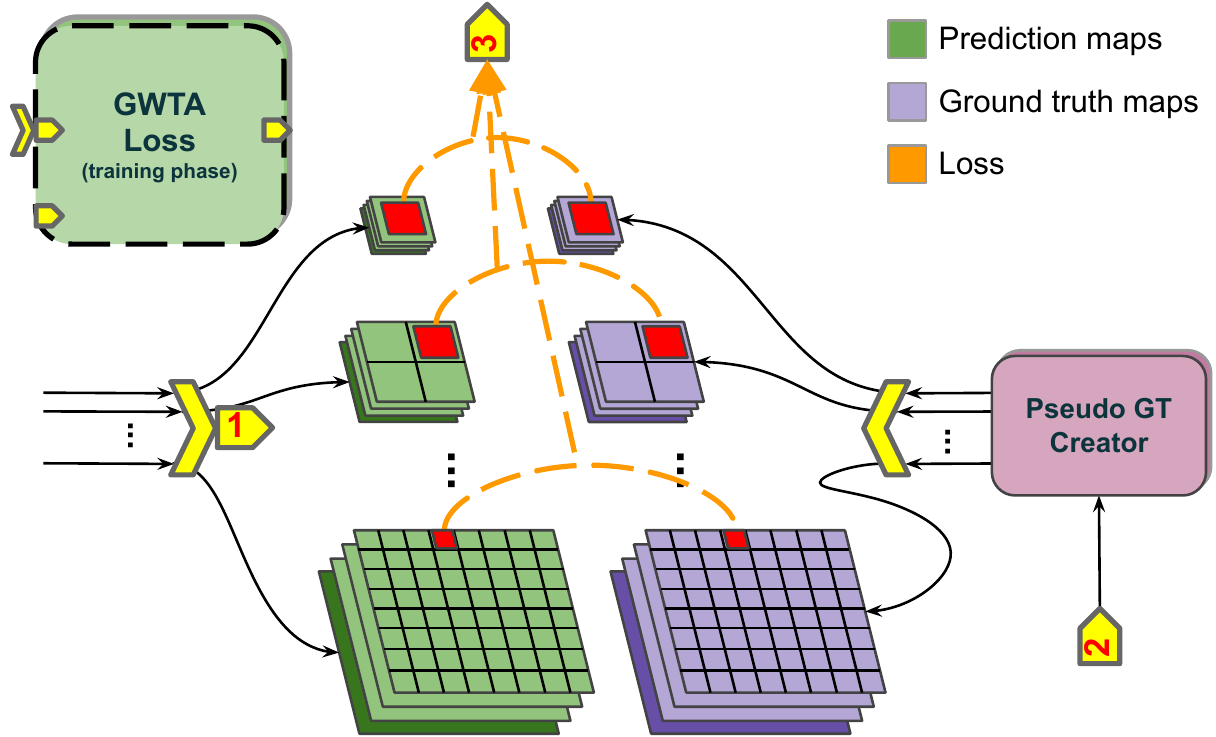}
    \end{center}
    \caption{Illustration of the operations in GWTA training. GWTA only selects the highest loss making cell in every scale. The per-pixel  cross-entropy loss is computed between the prediction and pseudo ground truth maps.}
    \label{fig:WTA}
\end{figure}

Figure \ref{fig:WTA} shows the implementation of GWTA training. For each scale, we apply GWTA non-linearity 
on the loss. The cell size for all branches is taken as the dimensions of the lowest resolution prediction 
map $(w^{0},h^{0})$. There is only one cell for scale $s=0$ (one-sixteenth branch), but the grows by power 
of four ($4^{s}$) for subsequent branches as the spatial dimensions consecutively doubles. Now we compute the 
cross-entropy loss for any cell at location $(x,y)$ (top-left corner) in the grid as, 
\small\[
l_{wta}^{s}[x,y]=\sum_{(\lfloor\frac{x'}{w^{0}}\rfloor,\lfloor\frac{y'}{h^{0}}\rfloor)=(x,y)}l(\{\mathcal{D}^{s}_{b}[x',y']\},\{\overline{\mathcal{B}}^{s}_{b}[x',y']\},\{\bar{\alpha}^{s}_{b}\}),
\]
where the summation of losses runs over for all pixels in the cell under consideration. Also note that 
$\bar{\alpha}^{s}_{b}$ is computed using equation \ref{equ:weights} with $c_{sum}^{s}=4^{-s}\sum_{b=1}^{n_{\mathcal{B}}}c^{s}_{b}$, in order to account for the change in spatial size 
of the predictions. The winner cell is the one with the highest loss and the location is given by,  
\begin{equation}
(x_{wta}^{s},y_{wta}^{s})=\underset{(x,y)=(w^{0}i,h^{0}j),i\in\mathbb{Z},j\in\mathbb{Z}}{\textrm{argmax}}l_{wta}^{s}[x,y].
\end{equation}
Note that the argmax operator finds an $(x,y)$ pair that identifies the top-left corner of the cell. 
The combined loss becomes,
\begin{equation}
\mathcal{L}_{wta}=\frac{1}{w^{0}h^{0}}\underset{s=1}{\stackrel{n_{\mathcal{S}}}{\sum}}l_{wta}^{s}[x_{wta}^{s},y_{wta}^{s}].
\end{equation}

\begin{figure*}[!t]
    \begin{center}
    \includegraphics[width=0.94\linewidth,height=0.378\textheight]{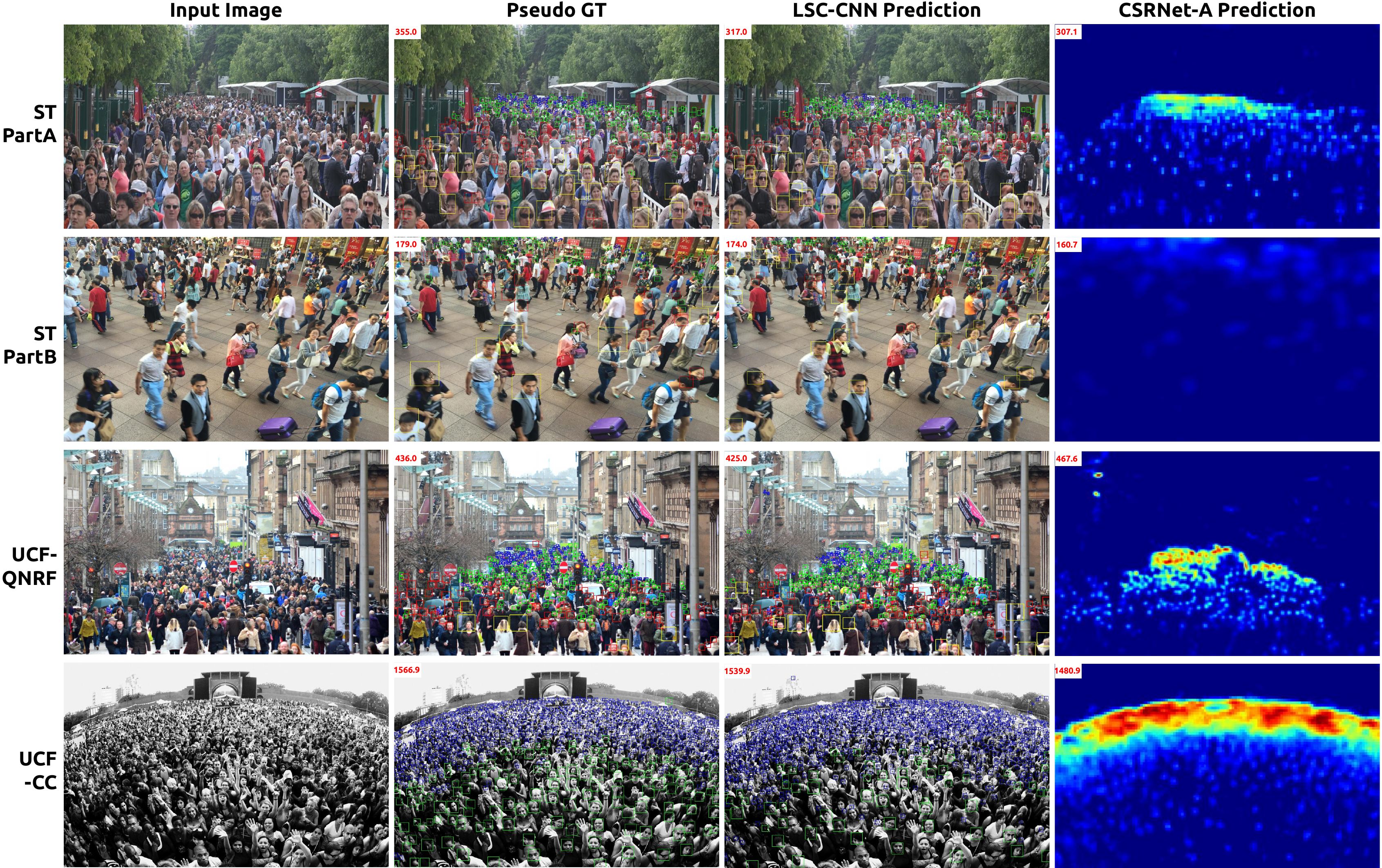}
    \end{center}
    \caption{Predictions made by LSC-CNN on images from Shanghaitech, UCF-QNRF and UCF-CC-50 datasets. The results emphasize the ability of our approach to pinpoint people consistently across crowds of different types than the baseline density regression method.}
    \label{fig:f2}
\end{figure*}

We optimize the parameters of LSC-CNN by backpropagating $\mathcal{L}_{wta}$ using standard mini-batch gradient descent 
with momentum. Batch size is typically 4. Momentum parameter is set as 0.9 and a fixed learning rate schedule of $10^{-3}$ 
is used. The training is continued till the counting performance (MAE in Section \ref{sect:localization}) on a validation 
set saturates.

\subsection{Count Heads}

\subsubsection{Prediction Fusion}

For testing the model, the GWTA training module is replaced with the \emph{prediction fusion} operation as shown 
in Figure \ref{fig:main-arch}. The input image is evaluated by all the branches and results in predictions at multiple 
resolutions. Box locations are extracted from these prediction maps and are linearly scaled to the input resolution. 
Then standard Non-Maximum Suppression (NMS) is applied to remove boxes with overlap more than a threshold. The boxes 
after the NMS form the final prediction of the model and are enumerated to output the crowd count. Note that, in order to 
facilitate intermediate evaluations during training, the NMS threshold is set to 0.3 (30\% area overlap). But for 
the best model after training, we run a threshold search to minimize the counting error (MAE, Section 
\ref{sect:localization}) over the validation set (typical value ranges from 0.2 to 0.3).

\section{Performance Evaluation}
\label{sect:performance-eval}

\subsection{Experimental Setup and Datasets\label{sect:expsetup}}

We evaluate LSC-CNN for localization and counting performance on all major crowd datasets. Since these 
datasets have only point head annotations, sizing capability cannot be benchmarked. Hence, we use one 
face detection dataset where bounding box ground truth is available. Further, LSC-CNN is trained on vehicle 
counting dataset to show generalization. Figure \ref{fig:f2} displays some of the box detections by our 
model on all datasets. Note that unless otherwise specified, we use the same architecture and 
hyper-parameters given in Section \ref{sect:approach}. The remaining part of this section introduces 
the datasets along with the hyper-parameters if there is any change.

\textbf{Shanghaitech:} The Shanghaitech (ST) dataset \cite{zhang2016single} consists of total 1,198 crowd 
images with more than 0.3 million head annotations. It is divided into two sets, namely, Part\_A and Part\_B. 
Part\_A has density variations ranging from 33 to 3139 people per image with average count being 501.4. 
In contrast, images in Part\_B are relatively less diverse and sparser with an average density of 123.6.

\textbf{UCF-QNRF:} Idrees et al. \cite{Idrees_2018_ECCV} introduce UCF-QNRF dataset and by large 
the biggest, with 1201 images for training and 334 images for testing. The 1.2 million head annotations 
come from diverse crowd images with density varying from as small as 49 people per image to massive 12865.

\textbf{UCF\_CC\_50:} UCF\_CC\_50 \cite{idrees2013multi} is a challenging dataset on multiple counts; the first is 
due to being a small set of 50 images and the second results from the extreme diversity, with crowd counts ranging 
in 50 to 4543. The small size poses a serious problem for training deep neural networks. Hence to 
reduce the number of parameters for training, we only use the one-eighth and one-fourth scale branches for this dataset. 
The prediction at one-sixteenth scale is avoided as sparse crowds are very less, but the top-down connections are kept as 
it is in Figure \ref{fig:main-arch}. The box increments are chosen as $\gamma=\{2,2\}$. Following 
\cite{idrees2013multi}, we perform 5-fold cross validation for evaluation.

\textbf{WorldExpo'10:} An array of 3980 frames collected from different video sequences of surveillance cameras forms 
the WorldExpo'10 dataset \cite{zhang2015cross}. It has sparse crowds with an average density of only 50 people per 
image. There are 3,380 images for training and 600 images from five different scenes form the test set. Region of 
Interest (RoI) masks are also provided for every scene. Since the dataset has mostly sparse crowds, only the 
low-resolution scales one-sixteenth and one-eighth are used with $\gamma=\{2,2\}$. 
We use a higher batch size of 32 as there many no people images and follow training/testing protocols in \cite{zhang2015cross}.

\textbf{TRANCOS:} The vehicle counting dataset, TRANCOS \cite{TRANCOSdataset_IbPRIA2015}, has 1244 images captured by 
various traffic surveillance cameras. In total, there are 46,796 vehicle point annotations. Also, RoIs are specified on 
every image for evaluation. We use the same architecture and box sizes as that of WorldExpo'10.

\textbf{WIDERFACE:} WIDERFACE \cite{widerface} is a face detection dataset with more than 0.3 million bounding box 
annotations, spanning 32,203 images. The images, in general, have sparse crowds having variations in pose and scale 
with some level of occlusions. We remove the one-half scale branch for this dataset as highly dense images are not 
present. To compare with existing methods on fitness of bounding box predictions, the fineness of the box sizes are 
increased by using five boxes per scale ($n_{\mathcal{B}}=5$). The $\gamma$ is set as $\{4,2,2\}$ and learning rate 
is made lower to $10^{-4}$. Note that for fair comparison, we train LSC-CNN without using the actual ground truth 
bounding boxes. Instead, point face annotations are created by taking centers of the boxes, from which pseudo ground 
truth is generated as per the training regime of LSC-CNN. But the performance is evaluated with the actual ground 
truth.

\begin{table*}[!t]
\begin{centering}
\begin{tabular}{|c|c|c|c|c|c|c|c|c|c|c|}
\hline Metric
 & \multicolumn{2}{c|}{MLE $\downarrow$} & \multicolumn{2}{c|}{GAME(0) $\downarrow$} & \multicolumn{2}{c|}{GAME(1) $\downarrow$} & \multicolumn{2}{c|}{GAME(2) $\downarrow$} & \multicolumn{2}{c|}{GAME(3) $\downarrow$}\tabularnewline
\cline{1-11} 
Dataset $\downarrow$ / Method$\rightarrow$ & CSR-A-thr & LSC-CNN & CSR-A & LSC-CNN & CSR-A & LSC-CNN & CSR-A & LSC-CNN & CSR-A & LSC-CNN\tabularnewline
\hline 
\hline 
ST Part\_A & 16.8 & 9.6 & 72.6 & 66.4 & 75.5 & 70.2 & 112.9 & 94.6 & 149.2 & 136.5\tabularnewline
\hline 
ST Part\_B & 12.28 & 9.0 & 11.5 & 8.1 & 13.1 & 9.6& 21.0& 17.4 & 28.9 & 26.5 \tabularnewline
\hline 
UCF\_QNRF & 14.2 & 8.6 & 155.8 & 120.5 & 157.2 & 125.8 & 186.7 & 159.9 & 219.3 & 206.0	\tabularnewline
\hline 
UCF\_CC\_50 & 14.3 & 9.7 & 282.9 & 225.6 &  326.3 & 227.4 & 369.0 & 306.8 & 425.8 & 390.0\tabularnewline
\hline 
\end{tabular}\medskip{}

\par\end{centering}

\caption{Comparison of LSC-CNN on localization metrics against the baseline regression method. Our model seems to pinpoint persons more accurately.}
\label{tbl:t1}
\end{table*}

\subsection{Evaluation of Localization}
\label{sect:localization}

The widely used metric for crowd counting is the Mean Absolute Error or MAE. MAE is simply the 
absolute difference between the predicted and actual crowd counts averaged over all the images in 
the test set. Mathematically, $\textrm{MAE}=\frac{1}{N}\sum_{n=1}^{N}|C_{n}-C_{n}^{GT}|,$
where $C_{n}$ is the count predicted for input image $n$ for which the ground truth is $C_{n}^{GT}$.
The counting performance of a model is directly evident from the MAE value. Further to estimate 
the variance and hence robustness of the count prediction, Mean Squared Error or MSE is used. 
It is given by $\textrm{MSE}=\sqrt{\frac{1}{N}\sum_{n=1}^{N}(C_{n}-C_{n}^{GT})^{2}}.$ Though 
these metrics measure the accuracy of overall count prediction, localization of the predictions 
is not very evident. Hence, apart from standard MAE, we evaluate the ability of LSC-CNN to accurately 
pinpoint individual persons. An existing metric called Grid Average Mean absolute Error or GAME 
\cite{TRANCOSdataset_IbPRIA2015}, can roughly indicate coarse localization of count predictions. To 
compute GAME, the prediction map is divided into a grid of cells and the absolute count errors within 
each cell are averaged over grid. Table \ref{tbl:t1} compares the GAME values of LSC-CNN against a regression 
baseline model for different grid sizes. Note that GAME with only one cell, GAME(0), is same as MAE. We take 
the baseline as CSRNet-A \cite{Li_2018_CVPR} (labeled CSR-A) model as it has similarity to the 
\emph{Feature Extractor} and delivers near state-of-the-art results. Clearly, LSC-CNN has superior count 
localization than the density regression based CSR-A. 

One could also measure localization in terms of how close the prediction matches with ground truth point 
annotation. For this, we define a metric named Mean Localization Error (MLE), which computes the distance 
in pixels between the predicted person location to its ground truth averaged over test set. The predictions 
are matched to head annotations in a one-to-one fashion and a fixed penalty of 16 pixels is added for absent 
or spurious detections. Since CSR-A or any other density regression based counting models do not individually 
locate persons, we apply threshold on the density map to get detections (CSR-A-thr). But it is difficult to threshold 
density maps without loss of counting accuracy. We choose a threshold such that the resultant MAE is minimum 
over validation set. For CSR-A, the best thresholded MAE comes to be 167.1, instead of the original 72.6. As 
expected, MLE scores for LSC-CNN is significantly better than CSR-A, indicating sharp localization capacity.

\begin{table}[!t]
\begin{centering}
\begin{tabular}{|c|c|c|c|}
\hline 
Method & Easy & Medium & Hard\tabularnewline
\hline 
\hline
Faceness \cite{Yang_2015_ICCV} & 71.3 & 53.4 & 34.5\tabularnewline
\hline
Two Stage CNN \cite{widerface} & 68.1 & 61.4 & 32.3\tabularnewline
\hline
TinyFace \cite{Hu_2017_CVPR} & 92.5 & 91.0 & 80.6\tabularnewline
\hline
SSH \cite{Najibi_2017_ICCV} & 93.1 & 92.1 & 84.5\tabularnewline
\hline
\hline 
CSR-A-thr (baseline) & 30.2 & 41.9 & 33.5 \tabularnewline
\hline 
PSDNN \cite{liu2019point} & 60.5 & 60.5 & 39.6\tabularnewline
\hline
LSC-CNN (Pseudo GT) & 40.5 & 62.1 & 46.2\tabularnewline
\hline 
LSC-CNN (Actual GT) & 57.31 & 70.10 & 68.9 \tabularnewline
\hline 
\end{tabular}\medskip{}

\par\end{centering}

\caption{Evaluation of LSC-CNN box prediction on WIDERFACE \cite{widerface}. Our model and PSDNN are trained on pseudo ground truths, while others use full supervision. LSC-CNN has impressive mAP in \emph{Medium} and \emph{Hard} sets.}
\label{tbl:sizing_widerface}
\end{table}

\subsection{Evaluation of Sizing}

We follow other face detection works 
(\hspace{1sp}\cite{Najibi_2017_ICCV,Hu_2017_CVPR}) and use the standard mean 
Average Precision or mAP metric to assess the sizing ability of our model. For this, LSC-CNN is trained 
on WIDERFACE face dataset without the actual box ground truth as mentioned in Section \ref{sect:expsetup}. 
Table \ref{tbl:sizing_widerface} reports the comparison of mAP scores obtained by our model against other works. 
Despite using pseudo ground truth for training, LSC-CNN achieves a competitive performance, especially on 
\emph{Hard} and \emph{Medium} test sets, against the methods that use full box supervision. 
For baseline, we consider the CSR-A-thr model (Section \ref{sect:localization}) where the 
density outputs are processed to get head locations. These are subsequently converted to bounding boxes using 
the pseudo box algorithm of LSC-CNN and mAP scores are computed (\emph{CSR-A-thr (baseline)}). LSC-CNN 
beats the baseline by a strong margin, evidencing the superiority of the proposed box classification training.
We also compare 
with PSDNN model \cite{liu2019point} which trains on pseudo box ground truth similar to our model. Interestingly, 
LSC-CNN has higher mAP in the two difficult sets than that of PSDNN. Note that the images in \emph{Easy} set 
are mostly of very sparse crowds with faces appearing large. We lose out in mAP mainly due to the high 
discretization of box sizes on large faces. This is not unexpected as LSC-CNN is designed for dense 
crowds without bounding box annotations. But the fact that it works well on the relatively denser other two test sets, clearly shows the 
effectiveness of our proposed framework. 
For completeness, we train LSC-CNN with boxes generated 
from actual box annotations instead of the head locations (\emph{LSC-CNN (Actual GT)}). As expected LSC-CNN performance improved with the use of real box size data.

\begin{table}[t]
\begin{centering}
\begin{tabular}{|c|c|c|}
\hline 
Method & MAE & MSE \tabularnewline
\hline 
\hline   
Idrees et al. \cite{idrees2013multi} & 315 & 508 \tabularnewline
\hline 
MCNN \cite{zhang2016single} & 277 & 426 \tabularnewline
\hline 
CMTL \cite{sindagi2017cnn} & 252 & 514 \tabularnewline
\hline
SCNN \cite{sam2017switching} & 228 & 445 \tabularnewline
\hline 
Idrees et al. \cite{Idrees_2018_ECCV} & 132 & 191 \tabularnewline
\hline 
\hline 
LSC-CNN (Ours) & \textbf{120.5} & 218.2 \tabularnewline
\hline 
\end{tabular}\medskip{}

\par\end{centering}

\caption{Counting performance comparison of LSC-CNN on UCF-QNRF \cite{idrees2013multi}.}
\label{tbl:QNRF}
\end{table}

\begin{table}[t]
\begin{centering}
\begin{tabular}{|c|c|c|c|c|c|c|c|c|}
\hline 
 & \multicolumn{2}{c|}{ST Part\_A} & \multicolumn{2}{c|}{ST Part\_B} & \multicolumn{2}{c|}{UCF\_CC\_50}\tabularnewline
\cline{1-7} 
Models & MAE & MSE & MAE & MSE & MAE & MSE\tabularnewline
\hline 
\hline 
Zhang et al. \cite{zhang2015cross} & 181.8 & 277.7 & 32.0 & 49.8 & 467.0 & 498.5 \tabularnewline
\hline 
MCNN \cite{zhang2016single} & 110.2 & 173.2 & 26.4 & 41.3 & 377.6 & 509.1\tabularnewline
\hline 
SCNN \cite{sam2017switching} & 90.4 & 135.0 & 21.6 & 33.4 & 318.1 & 439.2\tabularnewline
\hline 
CP-CNN \cite{sindagi2017generating} & 73.6 & 106.4 & 20.1 & 30.1 & 295.8 & 320.9\tabularnewline
\hline 
IG-CNN \cite{Sam_2018_CVPR} & 72.5 & 118.2 & 13.6 & 21.1 & 291.4 & 349.4\tabularnewline
\hline 
Liu et al. \cite{unlabelled_liu_tpami} & 72.0 & 106.6 & 14.4 & 23.8 & 279.6 & 388.9\tabularnewline
\hline 
IC-CNN \cite{Ranjan_2018_ECCV} & 68.5 & 116.2 & 10.7 & 16.0 & 260.9 & 365.5\tabularnewline
\hline 
CSR-Net \cite{Li_2018_CVPR} &   68.2 & 115.0 & 10.6 & 16.0 & 266.1 & 397.5\tabularnewline
\hline 
SA-Net \cite{Cao_2018_ECCV} & 67.0 & \textbf{104.5} & 8.4 & 13.6 & 258.4 & 334.9\tabularnewline
\hline 
PSDNN\cite{liu2019point} & \textbf{65.9} & 112.3 & 9.1 & 14.2 & 359.4 & 514.8\tabularnewline
\hline
LSC-CNN & 66.4 & 117.0 & \textbf{8.1} & \textbf{12.7} & \textbf{225.6} & \textbf{302.7}\tabularnewline
\hline 
\end{tabular}\medskip{}
\par\end{centering}

\caption{Benchmarking LSC-CNN counting accuracy on Shanghaitech \cite{zhang2016single} and UCF\_CC\_50 \cite{idrees2013multi} datasets. LSC-CNN stands state-of-the-art in both ST PartB and UCF\_CC\_50, with very competitive MAE on ST PartA.}
\label{tbl:counting-metrics}
\end{table}

We also compute the average classification accuracy of boxes with respect to the pseudo ground truth on test 
set. LSC-CNN has an accuracy of around 94.56\% for ST\_PartA dataset and 93.97\% for UCF\_QNRF, indicative of  
proper data fitting.

\subsection{Evaluation of Counting}
\label{sect:counting}

Here we compare LSC-CNN with other crowd counting models on the standard MAE and MSE metrics. Table 
\ref{tbl:QNRF} lists the evaluation results on UCF-QNRF dataset. Our model achieves an MAE of 
120.5, which is lower than that of \cite{Idrees_2018_ECCV} by a significant margin of 12.5. Evaluation 
on the next set of datasets is available in Table \ref{tbl:counting-metrics}. On Part\_A of Shanghaitech, 
LSC-CNN performs better than all the other density regression methods and has very competitive MAE to that 
of PSDNN \cite{liu2019point}, with the difference being just 0.5. But note that PSDNN is trained 
with a curriculum learning strategy and the MAE without it seems to be significantly higher (above 80). 
This along with the fact that LSC-CNN has lower count error than PSDNN in all other datasets, indicates 
the strength of our proposed architecture. In fact, state-of-the-art performance is obtained in 
both Shanghaitech Part\_B and UCF\_CC\_50 datasets. Despite having just 50 images with extreme diversity 
in the UCF\_CC\_50, our model delivers a substantial decrease of ~33 points in MAE. A similar 
trend is observed in WorldExpo dataset as well, with LSC-CNN acheiving lower MAE than existing methods 
(Table \ref{tbl:we}). Further to explore the generalization of LSC-CNN, we evaluate on a vehicle counting 
dataset TRANCOS. The results from Table \ref{tbl:TRANCOS} evidence a lower MAE than PSDNN, and is highly 
competitive with the best method. These experiments evidence the top-notch crowd counting 
ability of LSC-CNN compared to other density regressors, with all the merits of a detection model.

\begin{table}[t]
\begin{centering}
\begin{tabular}{|c|c|c|c|c|c|c|}
\hline 
Method & S1 & S2 & S3 & S4 & S5 & Avg.\tabularnewline
\hline 
\hline 
Zhang et al. \cite{zhang2015cross} & 9.8 & 14.1 & 14.3 & 22.2 & 3.7 & 12.9\tabularnewline
\hline 
MCNN \cite{zhang2016single} & 3.4 & 20.6 & 12.9 & 13.0 & 8.1 & 11.6\tabularnewline
\hline 
SCNN \cite{sam2017switching} & 4.4 & 15.7 & 10.0 & 11.0 & 5.9 & 9.4\tabularnewline
\hline 
CP-CNN \cite{sindagi2017generating} & 2.9 & 14.7 & 10.5 & 10.4 & 5.8 & 8.8\tabularnewline
\hline 
Liu et al. \cite{unlabelled_liu_tpami} & 2.0 & 13.1 & 8.9 & 17.4 & 4.8 & 9.2\tabularnewline
\hline 
IC-CNN \cite{Ranjan_2018_ECCV} & 17.0 & 12.3 & 9.2 & 8.1 & 4.7 & 10.3\tabularnewline
\hline
CSR-Net \cite{Li_2018_CVPR} & 2.9 & 11.5 & 8.6 & 16.6 & 3.4 & 8.6\tabularnewline
\hline
SA-Net \cite{Cao_2018_ECCV} & 2.6 & 13.2 & 9.0 & 13.3 & 3.0 & 8.2\tabularnewline
\hline 
\hline 
LSC-CNN (Ours) & 2.9 & 11.3 & 9.4 & 12.3 & 4.3 & \textbf{8.0}\tabularnewline
\hline
\end{tabular}\medskip{}
\par\end{centering}

\caption{LSC-CNN on WorldExpo'10 \cite{idrees2013multi} beats other methods in average MAE.}
\label{tbl:we}
\end{table}

\begin{table}[!t]
\begin{centering}
\begin{tabular}{|c|c|c|c|c|}
\hline 
Method & GAME0 & GAME1 & GAME2 & GAME3\tabularnewline
\hline 
\hline
Guerrero et al. \cite{TRANCOSdataset_IbPRIA2015} & 14.0 & 18.1& 23.7& 28.4 \tabularnewline
\hline
Hydra CNN \cite{onoro2016towards} & 10.9 &13.8 &16.0 & 19.3\tabularnewline
\hline
Li et al. \cite{Li_2018_CVPR} & \textbf{3.7} & 5.5 & 8.6& 15.0 \tabularnewline
\hline 
PSDNN \cite{liu2019point} & 4.8 & \textbf{5.4} & \textbf{6.7} & 8.4 \tabularnewline
\hline
LSC-CNN (Ours) & 4.6 & \textbf{5.4} & 6.9& \textbf{8.3}\tabularnewline
\hline 
\end{tabular}\medskip{}

\par\end{centering}

\caption{Evaluation of LSC-CNN on TRANCOS \cite{TRANCOSdataset_IbPRIA2015} vehicle counting dataset.}
\label{tbl:TRANCOS}
\end{table}

\section{Analysis and Ablations}

\subsection{Effect of Multi-Scale Box Classification}
\label{sect:s5_1}

As mentioned in Section \ref{sect:size-persons}, in general, we use 3 box sizes ($n_{\mathcal{B}}=3$) 
for each scale branch and employ 4 scales ($n_{\mathcal{S}}=4$). Here we ablate over the choice of 
$n_{\mathcal{B}}$ and $n_{\mathcal{S}}$. The results of the experiments are presented in Table 
\ref{tbl:hparam_ablation}. It is intuitive to expect higher counting accuracy with more number of 
scale branches (from $n_\mathcal{S}=1$ to $n_\mathcal{S}=4$) as people at all the scales are 
resolved better. Although this is true in theory, as the number of scales increase, so do the number 
of trainable parameters for the same amount of data. This might be the cause for slight increase in 
counting error for $n_\mathcal{S}=5$. Regarding the ablations on the number of boxes, we train LSC-CNN  
for $n_{\mathcal{B}}=1$ to $n_{\mathcal{B}}=4$ (maintaining the same size increments $\gamma$ as 
specified in Section \ref{sect:size-persons} for all). Initially, we observe a progressive gain in the 
counting accuracy till $n_{\mathcal{B}}=3$, but seems to saturate after that. This could be attributed 
to the decrease in training samples per box class as $n_{\mathcal{B}}$ increases.

\subsection{Architectural Ablations}
\label{sect:arch-ablations}

\begin{table}[t]
\begin{centering}
\begin{tabular}{|c|c|c|c|}
\hline 
$n_{\mathcal{S}}$ & $n_{\mathcal{B}}$ & ST\_PartA \cite{zhang2016single} & UCF-QNRF \cite{Idrees_2018_ECCV}\tabularnewline
\hline 
\hline 
1 & 3 & 155.2 & 197.8\tabularnewline
\hline 
2 & 3 & 113.9 & 142.1\tabularnewline
\hline 
3 & 3 & 75.3 & 134.5\tabularnewline
\hline 
5 & 3 & 69.3 & 124.8\tabularnewline
\hline 
\hline 
4 & 1 & 104.7 & 145.6\tabularnewline
\hline 
4 & 2 & 72.6 & 132.3\tabularnewline
\hline 
4 & 4 & 74.3 & 125.4\tabularnewline
\hline 
\hline
4 & 3 & \textbf{66.4} & \textbf{120.5}\tabularnewline
\hline 
\end{tabular}\medskip{}
\par\end{centering}

\caption{MAE obtained by LSC-CNN with different hyper-parameter settings.}
\label{tbl:hparam_ablation}
\end{table}

\begin{table}[t]
\begin{centering}
\begin{tabular}{|c|c|c|c|c|c|}
\hline 
 & ST PartA & UCF-QNRF & \multicolumn{3}{c|}{WIDERFACE}\tabularnewline
\cline{4-6} 
Method & MAE & MAE & Easy & Med & Hard \tabularnewline
\hline 
\hline 
No TFM & 94.5 & 149.7 & 30.1 & 45.2 & 31.5\tabularnewline
\hline 
Seq TFM & 73.4 & 135.2 & 31.4 & 47.3 & 39.8\tabularnewline
\hline 
Mult TFM & 67.6 & 124.1 & 37.8 & 54.2 & 45.1\tabularnewline
\hline 
No GWTA & 79.2 & 130.2 & 31.7 & 49.9 & 37.2\tabularnewline
\hline 
No Weighing & 360.1 & 675.5 & 0.1 & 0.1 & 1.2\tabularnewline
\hline 
No Replication & 79.3 & 173.1 & 30.4 & 44.3 & 35.9\tabularnewline
\hline 
Box Regression & 77.9 & 140.6 & 29.8 & 47.8 & 35.2\tabularnewline
\hline 
LSC-CNN & \textbf{66.4} & \textbf{120.5} & 40.5 & 62.1 & 56.2\tabularnewline
\hline 
\end{tabular}\medskip{}
\par\end{centering}
\caption{Validating various architectural design choices of LSC-CNN.}
\label{tbl:arch_ablation}
\end{table}

\begin{figure}[b]
    \begin{center}
    \includegraphics[width=\linewidth]{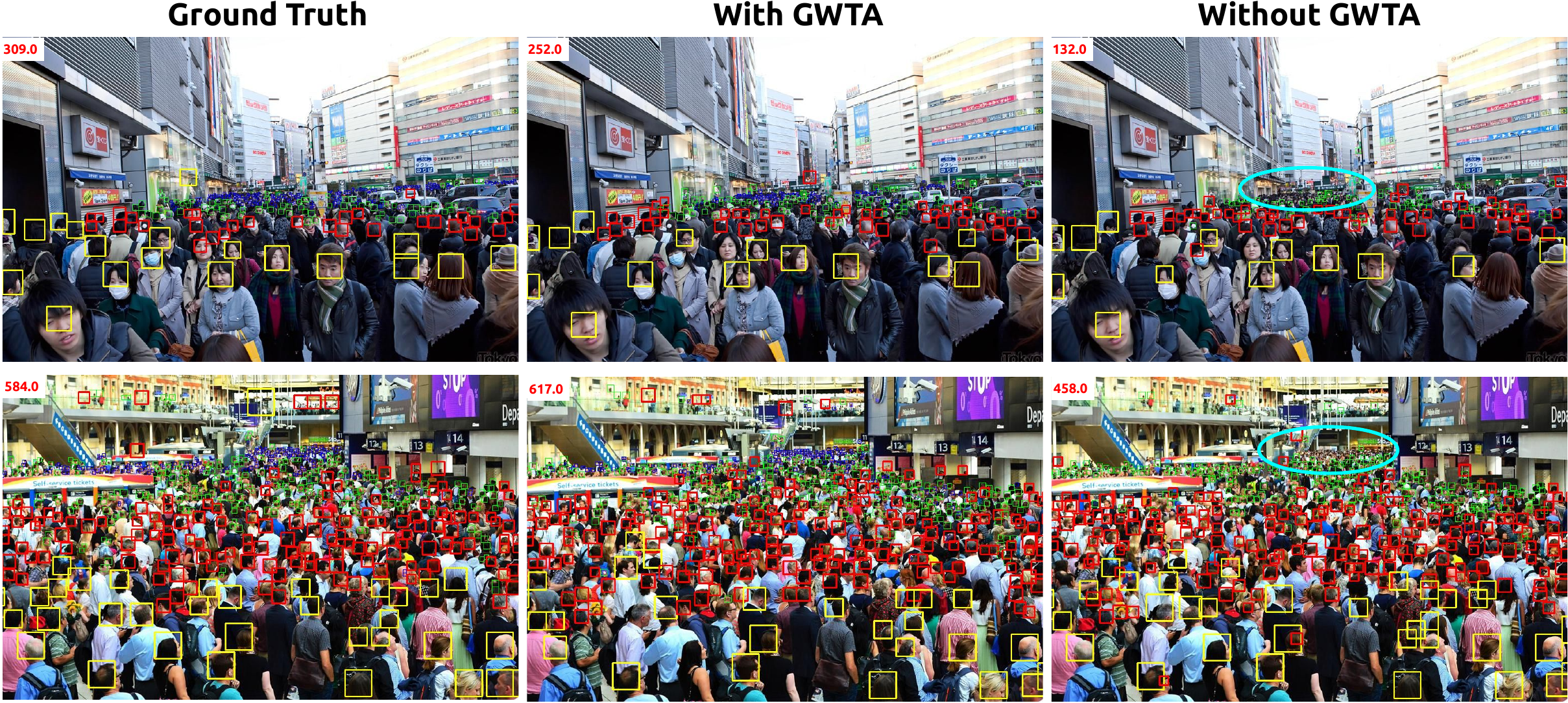}
    \end{center}
    \caption{Demonstrating the effectiveness of GWTA in proper training of high resolution scale branches (notice the highlighted region).}
    \label{fig:gtwa_comparison}
\end{figure}

\begin{figure*}[!t]
    \begin{center}
    \includegraphics[width=0.93\linewidth]{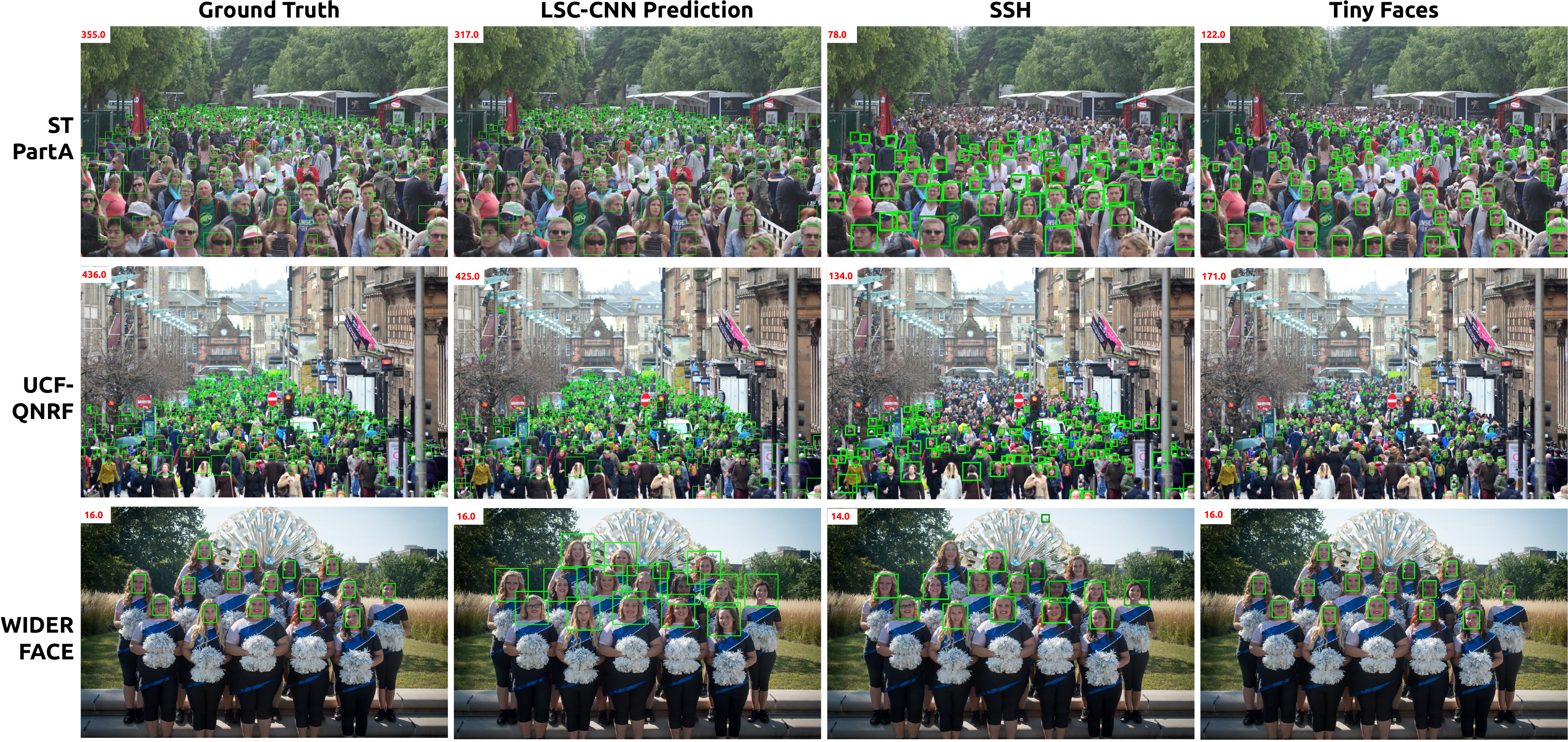}
    \end{center}
    \caption{Comparison of predictions made by face detectors SSH \cite{Najibi_2017_ICCV} and TinyFaces \cite{Hu_2017_CVPR} against LSC-CNN. Note that the \emph{Ground Truth} shown for WIDERFACE dataset is the actual and not the pseudo box ground truth. Normal face detectors are seen to fail on dense crowds.}
    \label{fig:detectors-comparison}
\end{figure*}

In this section, the advantage of various architectural choices made for our model is established through 
experiments. LSC-CNN employs multi-scale top-down modulation through the TFM modules (Section \ref{sect:TFM}). 
We train LSC-CNN without these top-down connections (terminal $3$ in Figure \ref{fig:TFM} is removed for all 
TFM networks) and the resultant MAE is labeled as \emph{No TFM} in Table \ref{tbl:arch_ablation}. We also 
ablate with a sequential TFM (\emph{Seq TFM}), in which every branch gets only one top-down connection from 
its previous scale as opposed to features from all lower resolution scales in LSC-CNN. The results evidence 
that having top-down modulation is effective in leveraging high-level scene context and helps improve count accuracy.
But the improvement is drastic with the proposed multiple top-down connections and seems to aid better 
extraction of context information. The top-down modulation can be incorporated in many ways, with LSC-CNN using 
concatenation of top-down features with that of bottom-up. Following \cite{AAAItdfcnn}, we generate features  
to gate the bottom-up feature maps (\emph{Mult TFM}). Specifically, we modify the second convolution layer 
for top-down processing in Figure \ref{fig:TFM} with Sigmoid activation. The Sigmoid output from each top-down connection  
is element-wise multiplied to the incoming scale feature maps. A slight performance drop is observed with 
this setup, but the MAE is close to that of LSC-CNN, stressing that top-down modulation in any form could be useful.

Now we ablate the training regime of LSC-CNN. The experiment labeled \emph{No GWTA} in Table 
\ref{tbl:arch_ablation} corresponds to LSC-CNN trained with just the $\mathcal{L}_{comb}$ loss (equation 
\ref{equ:lcomb}). 
Figure \ref{fig:gtwa_comparison} clearly shows that without GWTA, LSC-CNN completely fails in the 
high resolution scale (one-half), where the gradient averaging effect is prominent.
A significant drop in MAE is observed as well, validating the hypothesis 
that GWTA aids better optimization of the model. 
Another important aspect of the training is the class 
balancing scheme employed. LSC-CNN is trained with \emph{no weighting}, essentially with all $\alpha_{b}^{s}$s 
set to 1. As expected, the counting error reaches an unacceptable level, mainly due to the skewness in the 
distribution of persons across scales.
We also validate the usefulness of replicating certain VGG blocks in the 
\emph{feature extractor} (Section \ref{sect:feat_extr}) through an experiment without it, labeled as 
\emph{No Replication}.
Lastly, instead of our per-pixel box classification framework, we 
train LSC-CNN to regress the box sizes. \emph{Box regression} is done for all the branches by replicating 
the last five convolutional layers of the TFM (Figure \ref{fig:TFM}) into two arms; one for the per-pixel 
binary classification to locate person and the other for estimating the corresponding head sizes (the sizes 
are normalized to 0-1 for all scales). However, this setting could not achieve good MAE, possibly due to class imbalance across box sizes (Section \ref{sect:size-persons}).

\begin{table}[!t]
\begin{centering}
\begin{tabular}{|c|c|c|c|c|c|c|}
\hline 
 & \multicolumn{3}{c|}{ST Part\_A} & \multicolumn{3}{c|}{UCF-QNRF}\tabularnewline
\cline{1-7} 
Method & MAE & MSE & MLE & MAE & MSE & MLE\tabularnewline
\hline 
\hline 
FRCNN \cite{fasterrcnn} & 241.0 & 431.6 & 43.7 & 320.1 & 697.6 & 43.9\tabularnewline
\hline 
SSH (def) & 387.5 & 513.4 & 96.2 & 564.8 & 924.4  & 126.5 \tabularnewline
\hline 
SSH \cite{Najibi_2017_ICCV} & 328.2 & 479.6 & 89.9 & 441.1 & 796.6 & 103.7 \tabularnewline
\hline 
TinyFace (def) & 288.1 & 457.4  & 37.4 & 397.2 & 786.6 & 50.7\tabularnewline
\hline 
TinyFace \cite{Hu_2017_CVPR} & 237.8 & 422.8  & 29.6 & 336.8 & 741.6 & 41.2\tabularnewline
\hline 
LSC-CNN & \textbf{66.4} & \textbf{117.0} & \textbf{9.6} & \textbf{120.5} & \textbf{218.3} & \textbf{8.6}\tabularnewline
\hline 
\end{tabular}\medskip{}

\par\end{centering}

\caption{LSC-CNN compared with existing detectors trained on crowd datasets.}
\label{tbl:tbl7_det}
\end{table}

\subsection{Comparison with Object/Face Detectors}
\label{sect:comparision-with-detectors}

To further demonstrate the utility of our framework beyond any doubt, we train existing detectors like 
FRCNN~\cite{fasterrcnn}, SSH~\cite{Najibi_2017_ICCV} and TinyFaces~\cite{Hu_2017_CVPR} on dense crowd datasets. 
The anchors for these models are adjusted to match the box sizes ($\beta$s) of LSC-CNN for fair comparison. The 
models are optimized with the pseudo box ground truth generated from point annotations. 
For these, we compute counting metrics MAE and MSE along with point localization measure MLE in Table 
\ref{tbl:tbl7_det}. Note that the SSH and TinyFaces face detectors are also trained with the default anchor box 
setting as specified by their authors (labeled as \emph{def}). The evaluation points to the poor counting performance 
of the detectors, which incur high MAE scores. This is 
mainly due to the inability to capture dense crowds as evident from Figure \ref{fig:detectors-comparison}. LSC-CNN, on the other hand, works well across density ranges, with quite convincing 
detections even on sparse crowd images from WIDERFACE \cite{widerface}. 
In addition, we compare the detectors for per image inference time (averaged over ST Part\_A \cite{zhang2016single} test set, evaluated on a NVIDIA V100 GPU) and model size in Table \ref{tbl:detector_efficiency}. The results reiterate the suitability of LSC-CNN for practical applications.

\begin{table}[!t]
\begin{centering}
\begin{tabular}{|c|c|c|c|c|}
\hline 
Method & Inference Time (ms) & Parameters (in millions)\tabularnewline
\hline 
\hline
FRCNN \cite{fasterrcnn} & 231.4 & 41.5\tabularnewline
\hline
SSH \cite{Najibi_2017_ICCV} & 48.1 & 19.8\tabularnewline
\hline
TinyFace \cite{Hu_2017_CVPR} & 348.6 & 30.0\tabularnewline
\hline 
LSC-CNN $n_{\mathcal{S}}=1$ & 29.4 & 12.9\tabularnewline
\hline 
LSC-CNN $n_{\mathcal{S}}=2$ & 32.3 & 18.3\tabularnewline
\hline 
LSC-CNN $n_{\mathcal{S}}=3$ & 50.6 & 20.6\tabularnewline
\hline 
LSC-CNN $n_{\mathcal{S}}=4$ & 69.0 & 21.9\tabularnewline
\hline 
\end{tabular}\medskip{}

\par\end{centering}

\caption{Efficiency of detectors in terms of inference speed and model size.}
\label{tbl:detector_efficiency}
\end{table}

\section{Conclusion}
This paper introduces a \emph{dense detection} framework for crowd counting 
and renders the prevalent paradigm of density regression obsolete. The proposed 
LSC-CNN model uses a multi-column architecture with top-down modulation  
to resolve people in dense crowds. Though only point head annotations are available 
for training, LSC-CNN puts bounding box on every located person. Experiments indicate that the model achieves not only better crowd counting 
performance than existing regression methods, but also has superior localization 
with all the merits of a detection system. Given these, we hope that the community would switch 
from the current regression approach to more practical \emph{dense detection}.
Future research could address spurious detections and 
make sizing of heads further accurate.



\ifCLASSOPTIONcompsoc
  \section*{Acknowledgments}
\else
  \section*{Acknowledgment}
\fi
This work was supported by SERB, Dept. of Science and Technology, Govt. of India (Proj: SB/S3/EECE/0127/2015).

\ifCLASSOPTIONcaptionsoff
  \newpage
\fi



\bibliographystyle{IEEEtran}
\bibliography{IEEEabrv,references.bib}

\begin{thebibliography}{10}
\providecommand{\url}[1]{#1}
\csname url@samestyle\endcsname
\providecommand{\newblock}{\relax}
\providecommand{\bibinfo}[2]{#2}
\providecommand{\BIBentrySTDinterwordspacing}{\spaceskip=0pt\relax}
\providecommand{\BIBentryALTinterwordstretchfactor}{4}
\providecommand{\BIBentryALTinterwordspacing}{\spaceskip=\fontdimen2\font plus
\BIBentryALTinterwordstretchfactor\fontdimen3\font minus
  \fontdimen4\font\relax}
\providecommand{\BIBforeignlanguage}[2]{{%
\expandafter\ifx\csname l@#1\endcsname\relax
\typeout{** WARNING: IEEEtran.bst: No hyphenation pattern has been}%
\typeout{** loaded for the language `#1'. Using the pattern for}%
\typeout{** the default language instead.}%
\else
\language=\csname l@#1\endcsname
\fi
#2}}
\providecommand{\BIBdecl}{\relax}
\BIBdecl

\bibitem{Hu_2017_CVPR}
P.~Hu and D.~Ramanan, ``Finding tiny faces,'' in \emph{Proceedings of the IEEE
  Conference on Computer Vision and Pattern Recognition}, 2017.

\bibitem{widerface}
S.~Yang, P.~Luo, C.~C. Loy, and X.~Tang, ``{WIDER FACE}: A face detection
  benchmark,'' in \emph{Proceedings of the IEEE Conference on Computer Vision
  and Pattern Recognition (CVPR)}, 2016.

\bibitem{largestselfi}
``World's largest selfie,''
  \url{https://www.gsmarena.com/nokia_lumia_730_captures_worlds_largest_selfie-news-10285.php},
  accessed: 2019-05-31.

\bibitem{zhang2016single}
Y.~Zhang, D.~Zhou, S.~Chen, S.~Gao, and Y.~Ma, ``Single-image crowd counting
  via multi-column convolutional neural network,'' in \emph{Proceedings of the
  IEEE Conference on Computer Vision and Pattern Recognition (CVPR)}, 2016.

\bibitem{wu2005detection}
B.~Wu and R.~Nevatia, ``Detection of multiple, partially occluded humans in a
  single image by bayesian combination of edgelet part detectors,'' in
  \emph{Proceedings of the IEEE International Conference on Computer Vision
  (ICCV)}, 2005.

\bibitem{viola2005detecting}
P.~Viola, M.~J. Jones, and D.~Snow, ``Detecting pedestrians using patterns of
  motion and appearance,'' \emph{International Journal of Computer Vision
  (IJCV)}, 2005.

\bibitem{wang2011automatic}
M.~Wang and X.~Wang, ``Automatic adaptation of a generic pedestrian detector to
  a specific traffic scene,'' in \emph{Proceedings of the IEEE Conference on
  Computer Vision and Pattern Recognition (CVPR)}, 2011.

\bibitem{Idrees_tpami}
H.~{Idrees}, K.~{Soomro}, and M.~{Shah}, ``Detecting humans in dense crowds
  using locally-consistent scale prior and global occlusion reasoning,''
  \emph{IEEE Transactions on Pattern Analysis and Machine Intelligence
  (TPAMI)}, 2015.

\bibitem{idrees2013multi}
H.~Idrees, I.~Saleemi, C.~Seibert, and M.~Shah, ``Multi-source multi-scale
  counting in extremely dense crowd images,'' in \emph{Proceedings of the IEEE
  Conference on Computer Vision and Pattern Recognition (CVPR)}, 2013.

\bibitem{zhang2015cross}
C.~Zhang, H.~Li, X.~Wang, and X.~Yang, ``Cross-scene crowd counting via deep
  convolutional neural networks,'' in \emph{Proceedings of the IEEE Conference
  on Computer Vision and Pattern Recognition (CVPR)}, 2015.

\bibitem{onoro2016towards}
D.~Onoro-Rubio and R.~J. L{\'o}pez-Sastre, ``Towards perspective-free object
  counting with deep learning,'' in \emph{Proceedings of the European
  Conference on Computer Vision (ECCV)}, 2016.

\bibitem{sam2017switching}
D.~Babu~Sam, S.~Surya, and R.~V. Babu, ``Switching convolutional neural network
  for crowd counting,'' in \emph{Proceedings of the IEEE Conference on Computer
  Vision and Pattern Recognition (CVPR)}, 2017.

\bibitem{sindagi2017generating}
V.~A. Sindagi and V.~M. Patel, ``Generating high-quality crowd density maps
  using contextual pyramid {CNN}s,'' in \emph{Proceedings of the IEEE
  International Conference on Computer Vision (ICCV)}, 2017.

\bibitem{Li_2018_CVPR}
Y.~Li, X.~Zhang, and D.~Chen, ``{CSRNet}: Dilated convolutional neural networks
  for understanding the highly congested scenes,'' in \emph{Proceedings of the
  IEEE Conference on Computer Vision and Pattern Recognition (CVPR)}, 2018.

\bibitem{fasterrcnn}
S.~Ren, K.~He, R.~Girshick, and J.~Sun, ``Faster {R-CNN}: Towards real-time
  object detection with region proposal networks,'' in \emph{Advances in Neural
  Information Processing Systems ({NIPS})}, 2015.

\bibitem{liu2016ssd}
W.~Liu, D.~Anguelov, D.~Erhan, C.~Szegedy, S.~Reed, C.-Y. Fu, and A.~C. Berg,
  ``{SSD}: Single shot multibox detector,'' in \emph{Proceedings of the
  European Conference on Computer Vision (ECCV)}, 2016.

\bibitem{Redmon_2017_CVPR}
J.~Redmon and A.~Farhadi, ``{YOLO9000}: Better, faster, stronger,'' in
  \emph{Proceedings of the IEEE Conference on Computer Vision and Pattern
  Recognition (CVPR)}, 2017.

\bibitem{Najibi_2017_ICCV}
M.~Najibi, P.~Samangouei, R.~Chellappa, and L.~S. Davis, ``{SSH}: Single stage
  headless face detector,'' in \emph{Proceedings of the IEEE International
  Conference on Computer Vision (ICCV)}, 2017.

\bibitem{stewart2015end}
R.~Stewart and M.~Andriluka, ``End-to-end people detection in crowded scenes,''
  \emph{arXiv preprint arXiv:1506.04878}, 2015.

\bibitem{wang2015deep}
C.~Wang, H.~Zhang, L.~Yang, S.~Liu, and X.~Cao, ``Deep people counting in
  extremely dense crowds,'' in \emph{Proceedings of the ACM International
  Conference on Multimedia (ACMMM)}, 2015.

\bibitem{walach2016learning}
E.~Walach and L.~Wolf, ``Learning to count with {CNN} boosting,'' in
  \emph{Proceedings of the European Conference on Computer Vision}, 2016.

\bibitem{boominathan2016crowdnet}
L.~Boominathan, S.~S. Kruthiventi, and R.~V. Babu, ``{CrowdNet}: A deep
  convolutional network for dense crowd counting,'' in \emph{Proceedings of the
  ACM International Conference on Multimedia ({ACMMM})}, 2016.

\bibitem{Sam_2018_CVPR}
D.~Babu~Sam, N.~N. Sajjan, R.~V. Babu, and M.~Srinivasan, ``Divide and grow:
  Capturing huge diversity in crowd images with incrementally growing {CNN},''
  in \emph{Proceedings of the IEEE Conference on Computer Vision and Pattern
  Recognition (CVPR)}, 2018.

\bibitem{Cao_2018_ECCV}
X.~Cao, Z.~Wang, Y.~Zhao, and F.~Su, ``Scale aggregation network for accurate
  and efficient crowd counting,'' in \emph{Proceedings of the European
  Conference on Computer Vision (ECCV)}, 2018.

\bibitem{sindagi2017cnn}
V.~A. Sindagi and V.~M. Patel, ``{CNN}-based cascaded multi-task learning of
  high-level prior and density estimation for crowd counting,'' in
  \emph{Proceedings of the IEEE International Conference on Advanced Video and
  Signal Based Surveillance (AVSS)}, 2017.

\bibitem{AAAItdfcnn}
D.~Babu~Sam and R.~V. Babu, ``Top-down feedback for crowd counting
  convolutional neural network,'' in \emph{Proceedings of the AAAI Conference
  on Artificial Intelligence}, 2018.

\bibitem{Ranjan_2018_ECCV}
V.~Ranjan, H.~Le, and M.~Hoai, ``Iterative crowd counting,'' in
  \emph{Proceedings of the European Conference on Computer Vision}, 2018.

\bibitem{Liu_2018_CVPR}
J.~Liu, C.~Gao, D.~Meng, and A.~G. Hauptmann, ``{DecideNet}: Counting varying
  density crowds through attention guided detection and density estimation,''
  in \emph{Proceedings of the IEEE Conference on Computer Vision and Pattern
  Recognition (CVPR)}, 2018.

\bibitem{huang2017densely}
G.~Huang, Z.~Liu, L.~Van Der~Maaten, and K.~Q. Weinberger, ``Densely connected
  convolutional networks,'' in \emph{Proceedings of the IEEE Conference on
  Computer Vision and Pattern Recognition}, 2017.

\bibitem{Idrees_2018_ECCV}
H.~Idrees, M.~Tayyab, K.~Athrey, D.~Zhang, S.~Al-Maadeed, N.~Rajpoot, and
  M.~Shah, ``Composition loss for counting, density map estimation and
  localization in dense crowds,'' in \emph{Proceedings of the European
  Conference on Computer Vision (ECCV)}, 2018.

\bibitem{unlabelled_liu_tpami}
X.~{Liu}, J.~{Van De Weijer}, and A.~D. {Bagdanov}, ``Exploiting unlabeled data
  in {CNN}s by self-supervised learning to rank,'' \emph{IEEE Transactions on
  Pattern Analysis and Machine Intelligence}, 2019.

\bibitem{almost_unsup}
D.~Babu~Sam, N.~N. Sajjan, H.~Maurya, and R.~V. Babu, ``Almost unsupervised
  learning for dense crowd counting,'' in \emph{Proceedings of the {AAAI}
  Conference on Artificial Intelligence}, 2019.

\bibitem{Shi_2018_CVPR}
Z.~Shi, L.~Zhang, Y.~Liu, X.~Cao, Y.~Ye, M.-M. Cheng, and G.~Zheng, ``Crowd
  counting with deep negative correlation learning,'' in \emph{Proceedings of
  the IEEE Conference on Computer Vision and Pattern Recognition (CVPR)}, 2018.

\bibitem{Shen_2018_CVPR}
Z.~Shen, Y.~Xu, B.~Ni, M.~Wang, J.~Hu, and X.~Yang, ``Crowd counting via
  adversarial cross-scale consistency pursuit,'' in \emph{Proceedings of the
  IEEE Conference on Computer Vision and Pattern Recognition (CVPR)}, 2018.

\bibitem{Lowe04distinctiveimage}
D.~G. Lowe, ``Distinctive image features from scale-invariant keypoints,''
  \emph{International Journal of Computer Vision (IJCV)}, 2004.

\bibitem{girshick2014rich}
R.~Girshick, J.~Donahue, T.~Darrell, and J.~Malik, ``Rich feature hierarchies
  for accurate object detection and semantic segmentation,'' in
  \emph{Proceedings of the IEEE Conference on Computer Vision and Pattern
  Recognition (CVPR)}, 2014.

\bibitem{he2015spatial}
K.~He, X.~Zhang, S.~Ren, and J.~Sun, ``Spatial pyramid pooling in deep
  convolutional networks for visual recognition,'' \emph{IEEE Transactions on
  Pattern Analysis and Machine Intelligence}, 2015.

\bibitem{girshick2015fast}
R.~Girshick, ``Fast {R-CNN},'' in \emph{Proceedings of the IEEE International
  Conference on Computer Vision (CVPR)}, 2015.

\bibitem{zhu2017cms}
C.~Zhu, Y.~Zheng, K.~Luu, and M.~Savvides, ``{CMS-RCNN}: Contextual multi-scale
  region-based {CNN} for unconstrained face detection,'' in \emph{Deep Learning
  for Biometrics}.\hskip 1em plus 0.5em minus 0.4em\relax Springer, 2017.

\bibitem{sindagi2019dafe}
V.~A. Sindagi and V.~M. Patel, ``{DAFE-FD}: Density aware feature enrichment
  for face detection,'' in \emph{Proceedings of the IEEE Winter Conference on
  Applications of Computer Vision (WACV)}, 2019.

\bibitem{liu2019point}
Y.~Liu, M.~Shi, Q.~Zhao, and X.~Wang, ``Point in, box out: Beyond counting
  persons in crowds,'' in \emph{Proceedings of the IEEE Conference on Computer
  Vision and Pattern Recognition (CVPR)}, 2019.

\bibitem{makhzani2015winner}
A.~Makhzani and B.~J. Frey, ``Winner-take-all autoencoders,'' in \emph{Advances
  in Neural Information Processing Systems (NIPS)}, 2015.

\bibitem{simonyan2014very}
K.~Simonyan and A.~Zisserman, ``Very deep convolutional networks for
  large-scale image recognition,'' \emph{arXiv preprint arXiv:1409.1556}, 2014.

\bibitem{ILSVRC15}
O.~Russakovsky, J.~Deng, H.~Su, J.~Krause, S.~Satheesh, S.~Ma, Z.~Huang,
  A.~Karpathy, A.~Khosla, M.~Bernstein, A.~C. Berg, and L.~Fei-Fei, ``{ImageNet
  Large Scale Visual Recognition Challenge},'' \emph{International Journal of
  Computer Vision (IJCV)}, 2015.

\bibitem{TRANCOSdataset_IbPRIA2015}
R.~Guerrero-Gómez-Olmedo, B.~Torre-Jiménez, R.~López-Sastre, S.~M. Bascón,
  and D.~Oñoro-Rubio, ``Extremely overlapping vehicle counting,'' in
  \emph{Proceedings of the Iberian Conference on Pattern Recognition and Image
  Analysis (IbPRIA)}, 2015.

\bibitem{Yang_2015_ICCV}
S.~Yang, P.~Luo, C.-C. Loy, and X.~Tang, ``From facial parts responses to face
  detection: A deep learning approach,'' in \emph{Proceedings of the IEEE
  International Conference on Computer Vision}, 2015.

\end{thebibliography}
%



%




\end{document}